\documentclass[10pt, conference, compsocconf]{IEEEtran}
\usepackage[numbers]{natbib}

\ifCLASSINFOpdf
  \usepackage[pdftex]{graphicx}
  \graphicspath{{../pdf/}{../jpeg/}}
  \DeclareGraphicsExtensions{.pdf,.jpeg,.png}
\else
  \usepackage[dvips]{graphicx}
  \graphicspath{{../eps/}}
  \DeclareGraphicsExtensions{.eps}
\fi
\usepackage[cmex10]{amsmath}
\usepackage{algorithm}
\usepackage[algo2e]{algorithm2e}
\usepackage[caption=false,font=footnotesize]{subfig}
\usepackage{subfloat}
\usepackage{enumerate}
\usepackage{amsmath}
\usepackage{amssymb}
\usepackage{multirow}
\usepackage[flushleft]{threeparttable}

\begin{document}

\title{Inconsistent Node Flattening for Improving Top-down Hierarchical Classification}

\author{\IEEEauthorblockN{Azad Naik and Huzefa Rangwala}
\IEEEauthorblockA{
George Mason University\\
Fairfax, VA, United States\\
Email: anaik3@gmu.edu, rangwala@cs.gmu.edu}
}
\maketitle

\begin{abstract}
Large-scale classification of data where classes are structurally organized in a hierarchy is an important area of research. Top-down approaches that exploit the hierarchy during the learning and prediction phase are efficient for large-scale hierarchical classification. However, accuracy of top-down approaches is poor due to error propagation $i.e.$, prediction errors made at higher levels in the hierarchy cannot be corrected at lower levels. One of the main reason behind errors at the higher levels is the presence of inconsistent nodes that are introduced due to the arbitrary process of creating these hierarchies by domain experts. In this paper, we propose two different data-driven approaches (local and global) for hierarchical structure modification that identifies and flattens inconsistent nodes present within the hierarchy. Our extensive empirical evaluation of the proposed approaches on several image and text datasets with varying distribution of features, classes and training instances per class shows improved classification performance over competing hierarchical modification approaches. Specifically, we see an improvement upto 7$\%$ in Macro-F1 score with our approach over best TD baseline. SOURCE CODE: {http://www.cs.gmu.edu/$\sim$mlbio/InconsistentNodeFlattening}
\end{abstract}

\IEEEpubidadjcol

\section{Introduction}
Hierarchies (taxonomies) are the most commonly used data structure for organizing large volume of datasets in various domains including bioinformatics\footnote{http://geneontology.org/} for organizing gene sequences, ImageNet\footnote{http://image-net.org/} for organizing images and Yahoo!\footnote{http://dir.yahoo.com/} web directories for organizing web documents. Given the  hierarchy of classes, the Hierarchical Classification (HC) problem deals with learning models by utilizing (or ignoring) the hierarchical structure to automatically classify unlabeled test instances (examples) into relevant classes (categories). Over the  years, HC has gained immense interest among researchers and is evident from the various large-scale online prediction challenges such as LSHTC\footnote{http://lshtc.iit.demokritos.gr/}, BioASQ\footnote{http://www.bioasq.org/} and ILSVRC\footnote{http://image-net.org/challenges/LSVRC/}.

In the past, various methods have been developed to improve the HC performance \cite{silla2011survey}. One of the simplest method is to learn binary one-versus-rest classifier for each of the leaf categories, ignoring the hierarchical relationships. This method is referred as flat classification. Other methods involve use of the hierarchies (see Section \ref{lit}) during the learning and prediction process. Hierarchies provide useful structural relationships (such as parent-child and siblings) among different classes that can be exploited for learning generalized classification models. Previously, researchers have demonstrated the usefulness of hierarchies for classification and have obtained promising results \cite{gopal2013recursive,cai2004hierarchical,koller1997hierarchically,mccallum1998improving,dumais2000hierarchical,sun2001hierarchical,xue2008deep}. However, in many situations hierarchies used for learning models are not consistent due to the presence of \emph{inconsistent} nodes (and links) resulting in excessive error propagation. As such, HC approaches are outperformed by the flat classifiers that completely ignore the hierarchy \cite{xiao2011hierarchical, zimek2010study}.

\begin{figure*}
	\center
        \subfloat[][{{Original Hierarchy} ($\mathcal{H}$)}]{
            \includegraphics[width=.325\linewidth,height=2.25cm]{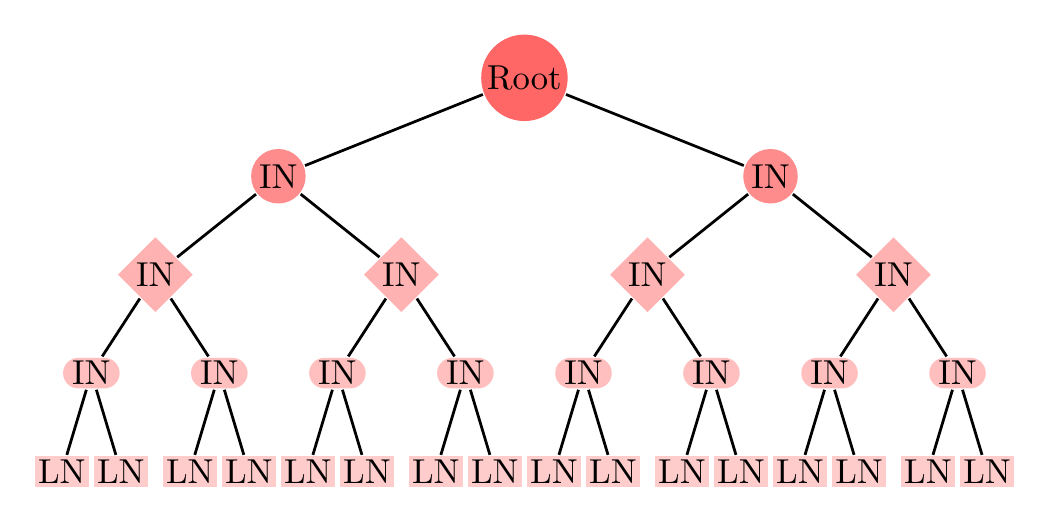}%
            \label{fig:initialhierarchy}}
        \hfil
        \subfloat[][{{Flat Hierarchy}}]{
            \includegraphics[width=.325\linewidth,height=2.25cm]{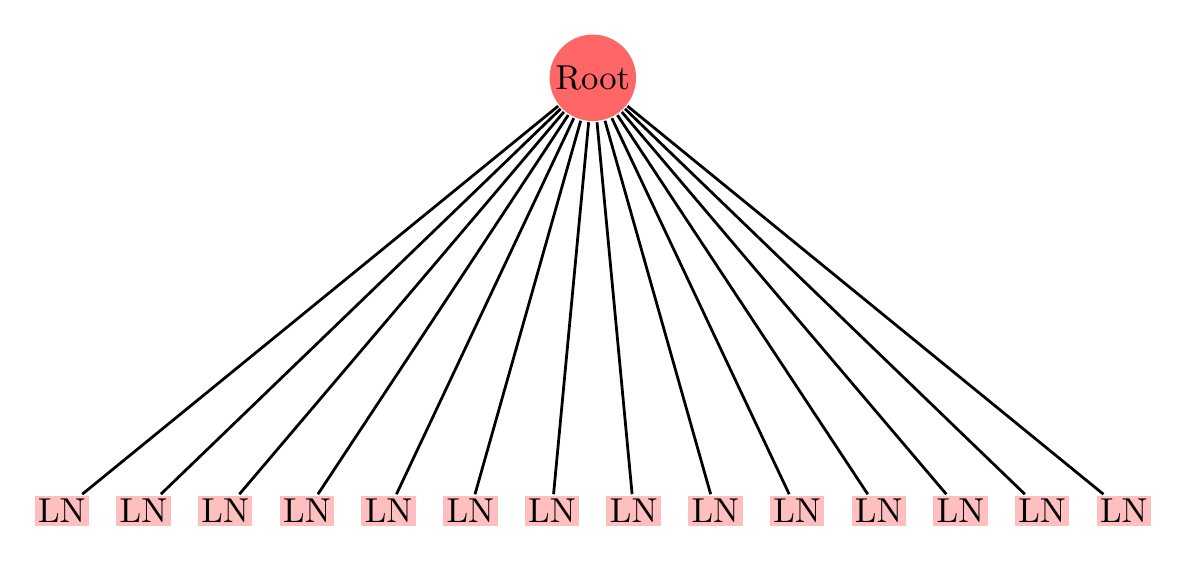}%
            \label{fig:fc}}
        \hfil
        \subfloat[][{{Top Level Flattened ({TLF}) Hierarchy}}]{
            \includegraphics[width=.325\linewidth,height=2.25cm]{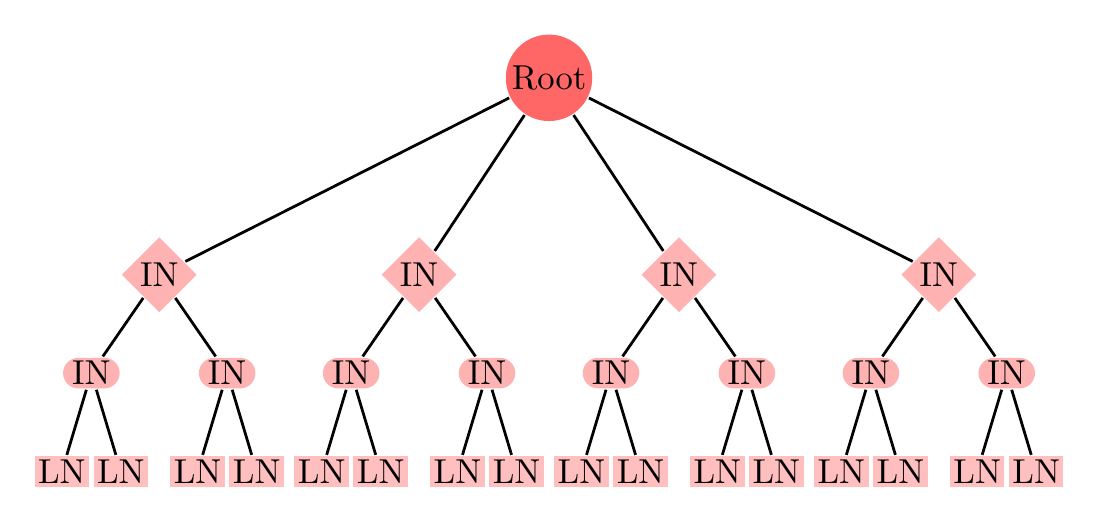}%
            \label{fig:tlf}}
        \hfil
        \subfloat[][{{Bottom Level Flattened ({BLF}) Hierarchy}}]{
            \includegraphics[width=.325\linewidth,height=2.25cm]{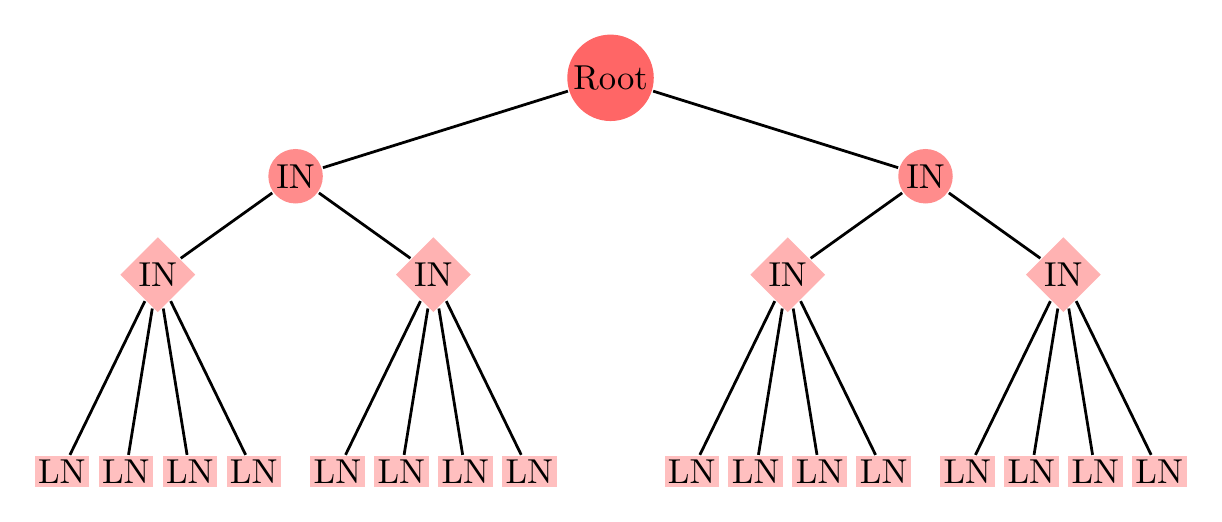}%
            \label{fig:blf}}%
        \hfil
        \subfloat[][{{Multiple Level Flattened ({MLF}) Hierarchy}}]{
            \includegraphics[width=.325\linewidth,height=2.25cm]{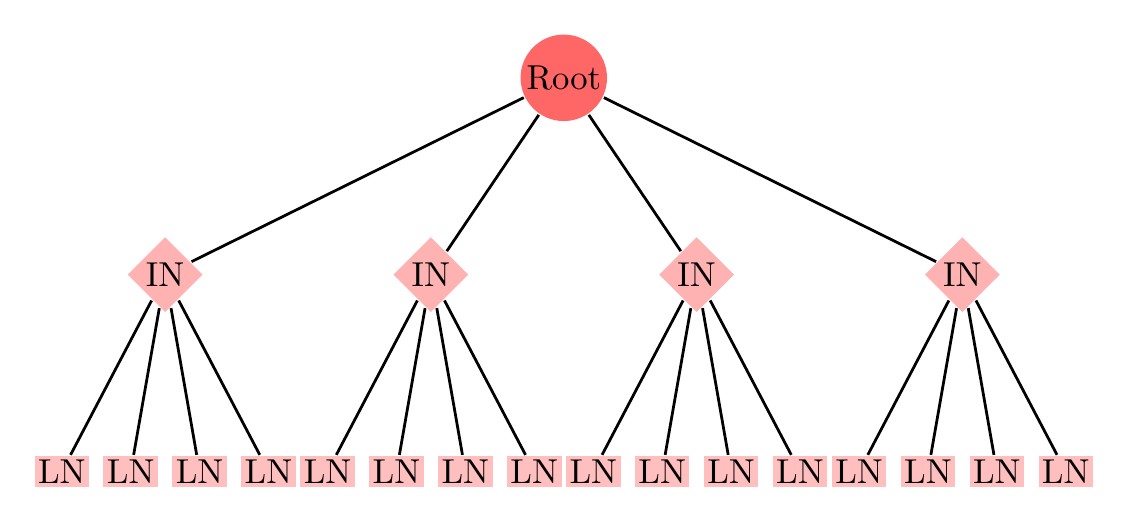}%
            \label{fig:mlf}}%
        \hfil
        \subfloat[][{{Inconsistent Node Flattened (INF) Hierarchy}}]{
            \includegraphics[width=.325\linewidth,height=2.25cm]{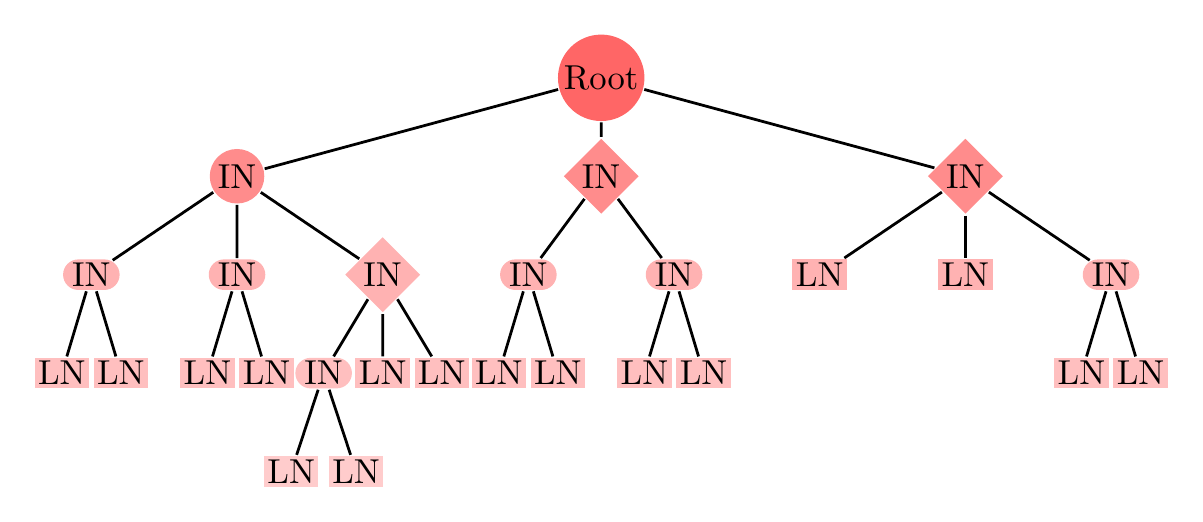}%
            \label{fig:linr}}%
    \caption{{Various hierarchical structures (b)-(f) obtained after flattening some of the nodes (or levels) from the original hierarchy shown in (a). `IN' denotes the internal node and `LN' denotes the leaf node.}}
    \label{hiefig}
    \vspace*{-5mm}
\end{figure*}

Flat classifiers,  though effective in some cases, suffer from two major issues: (i) During the prediction phase, flat classifiers invoke all the models for leaf categories and are considerably slower than top-down HC approaches, in which only the models in the relevant path are invoked. (ii) For large-scale HC problems, it is challenging to learn  effective classification models that can discriminate between large number of classes. This problem is worse for datasets with skewed class distributions where plenty of classes have very few examples for training (rare categories problem) \cite{babbar2013flat}. Large-scale datasets show 
power-law distribution of examples per category \cite{liu2005experimental}. Considering these issues, the  focus of  this paper is to improve top-down HC approaches, which are computationally feasible for large-scale datasets and handle
the imbalance problem by utilizing structural relationships.

The main drawback of top-down HC approaches that contributes to their inferior classification performance is error propagation --- compounding of errors from misclassifications at higher levels which cannot be rectified at the lower levels. This problem can be alleviated to certain extent by restructuring (modifying) the hierarchy to remove inconsistent nodes that causes performance deterioration. In this paper, our main contribution includes development of data-driven approaches for removing inconsistencies in the expert defined (original) hierarchy leading to a hierarchy that achieves higher classsification performance irrespective of the HC approach used for training. We propose a \emph{flattening}  approach where  inconsistent nodes 
are selectively removed from the hierarchy. The criterion for flattening a node is based on the optimal regularized risk minimization objective value attained by the model trained for that node on a separate validation set. If the objective value for a node $n$ is 
above a certain threshold, then we flatten $n$, \emph{i.e.}, we remove $n$ from the hierarchy and 
add its children to $n$'s parent node. Based on the strategy adapted for identifying inconsistent nodes, we propose two different approaches for inconsistent nodes flattening (INF) from the hierarchy: (i) Local approach (Level-INF) that computes a level-wise cutoff threshold and (ii) Global approach (Global-INF) that computes a global threshold for the entire hierarchy.

Experimental comparisons of top-down HC approach on our proposed modified hierarchy shows statistically significant performance improvement in comparison to the baseline hierarchy (expert defined or original) and other comparative methods for hierarchy modification \cite{wang2010flatten,babbar2013maximum}. We also performed detailed analysis to show that the reduction in misclassification error at higher levels with our proposed hierarchy modification approach leads to reduced error propagation and hence better classification performance. In comparison to flat classification, our approach is more accurate for classes with fewer training instances (rare categories) and is computationally efficient for large-scale HC problems during the prediction phase.

\section{Literature Review}
\label{lit}
There has been a large body of research focusing on the HC problem. Besides completely ignoring the hierarchy (flat classifiers), one class of HC methods solve various local subproblems that train individual classifier(s) for each of the nodes (or parent nodes) in the hierarchy or learn classifiers for each of the levels. This methods are referred as local classification because only  local structural relationship information are  used during training these 
classifiers. To predict the labels of instances, top-down local hierarchical methods proceed by selecting the most relevant node at the topmost level and then recursively selecting the best node until a leaf category is reached, which is the final predicted label. Local approaches are more popular due to their computational benefits \cite{silla2011survey}. Contrary to local classification, global classification methods \cite{vens2008decision} learn a single complex model over all the nodes in the hierarchy and are computationally more expensive than flat and local methods. Therefore, in this paper we focus on top-down local classification methods for training models and predicting labels.

Some of the earlier studies focus on exploiting hierarchies among categories for the purpose of classification \cite{cai2004hierarchical,dekel2009distribution,tang2006acclimatizing,dekel2004large}, but the number of categories are limited to a few hundreds. One of the earlier breakthroughs in the field of hierarchical text categorization was by Koller et al. \cite{koller1997hierarchically}. This approach used a divide and conquer paradigm for solving the HC problem which can easily be adopted in large-scale settings. Following this, numerous approaches have  been developed to improve HC for larger datasets. Liu et al. \cite{liu2005support} studied the classification performance using a SVM based method that scales for millions of categories. Gopal et al. \cite{gopal2013recursive} used a regularization term within the optimization function to capture the parent-child relationships in the hierarchy. This approach referred as HR-LR and HR-SVM shows improved classification performance but the training procedure  for large-scale problem requires a distributed implementation and map-reduce supported infrastructure. Xue et al. \cite{xue2008deep} proposed a two stage approach. For each test document, in the first stage a set of candidate categories are retrieved based on similarity to the test document. Then the second stage builds a classifier on the hierarchy restricted to the set of categories fetched in first stage and classifies the test document using the restricted hierarchy. Although, pruning reduces the hierarchy to a manageable size, one severe drawback of this approach is having to train a different classifier for each test document which is expensive. Other works in the field of HC can be found in a detailed survey by Silla et al. \cite{silla2011survey}. 

\subsection{Hierarchy Modification} 
\label{sec:lit:hm}
As discussed earlier, most HC approaches rely on hierachical relationships for learning complex  models to improve the classification performance. However, the performance can be negatively impacted if the hierarchy used during learning models is inconsistent. Therefore, it is of utmost importance to generate an improved hierarchical representation that is suitable for classification prior to learning models. Inconsistencies in the hierarchy are due to the following reasons: 
\begin{itemize}
\item [(i)] Hierarchies are designed for efficient search and navigation without considering HC performance. 
\item [(ii)] Hierarchical groupings of categories is done based on semantics, whereas classification depends on data characteristics such as \emph{term frequency}. 
\item [(iii)] Multiple hierarchies are possible for the same dataset (such as SCOP and CATH \cite{hadley1999systematic} for protein structures). However, there is no consensus regarding which hierarchy is better for classification. 
\item [(iv)] Consistent hierarchy design for datasets with large number of categories is prone to errors. 
\end{itemize}

Several approaches that restructure the hierarchy have been developed in past. Level flattening \cite{wang2010flatten} is one of the approach used in earlier works of hierarchy modification, where some of the levels are flattened (removed) from the original hierarchy prior to learning models.  Based on levels that are flattened various methods of modification exist. Top Level Flattening (TLF) as shown in Figure \ref{hiefig}(c) modifies the hierarchy by flattening the top level in the original hierarchy. Model learning and prediction for flattened hierarchy is done in similar fashion as top-down methods. Bottom Level Flattening (BLF) and Multiple Level Flattening (MLF), shown in Figures \ref{hiefig}(d) and \ref{hiefig}(e) are similar methods of hierarchy modification where bottom and multiple levels are removed, respectively. As done in Wang et al. \cite{wang2010flatten}, we removed the first and third levels for evaluating the MLF approach. 

Babbar et al. \cite{babbar2013maximum} proposed a maximum-margin based strategy for hierarchy modification. This method selectively removes some of the inconsistent nodes in the hierarchy based on margins rather than removing complete levels. Hierarchy modification using this approach (shown in Figure \ref{hiefig}(f)), is beneficial for classification and has been theoretically justified \cite{gao2011discriminative}. We followed a similar approach for hierarchy modification. However, our method differs in following two aspects: (i) We developed a more systematic approach for threshold selection to identify the inconsistent nodes for flattening. Our approach is based on deviation from mean that is empirically tuned for each dataset, and (ii) We also considered a global perspective of the hierarchy (Global-INF) for threshold selection to identify the inconsistent set of nodes. This approach is more intuitive and realistic measure for threshold selection because it prevents excessive flattening of the nodes that is just based on local decisions, thereby allowing the benefits of leveraging the hierarchy during the model learning and classification process, especially for rare categories (see Section \ref{resDis} for justification).

\begin{table}[t!] 
\centering 
\caption{\textbf{Notation description}} 
\label{table:notDes} 
\scriptsize
\begin{tabular}{|@{\hskip 0.015cm} l @{}| l @{}|} 
\hline
\multicolumn{1}{|c|}{\textbf{Symbol}} & \multicolumn{1}{c|}{\textbf{Description}}\\
\hline
$\mathcal{H}$ & original (experts defined) hierarchy\\
$\mathbb{R}$ & set of real numbers\\
$d$ & dimensionality of input vector\\
$N$ & total number of training examples\\
$\mathcal{L}$ & set of leaf categories\\
${\bf{x}}_i \in \mathbb{R}^d$ & input vector for $i$-th training example\\
$y_i \in \mathcal{L}$ & true label for $i$-th training example\\
$\hat{y}_i \in \mathcal{L}$ & predicted label for $i$-th test example\\
$y_i^n \in \pm 1$ & binary label used for $i$-th training example to learn weight vector for\\
& $n$-th node in $\mathcal{H}$, $y_i^n$ = 1 iff $y_i$ =$ n$, -1 otherwise\\
$\hat{y}_i^n \in \pm 1$ & predicted label for $i$-th test example at $n$-th node in the hierarchy, \\
& $\hat{y}_i^n$ = 1 iff prediction is positive, -1 otherwise\\
${\bf{\Theta}}_n$ & weight vector for $n$-th node\\
$C > 0$ & mis-classification penalty parameter\\
$f^*_n$ & optimal objective function value for $n$-th node obtained using  \\
& validation dataset. We have dropped the subscript $n$ at some places\\
& for ease of description\\
$\mathcal{H}_L$ & modified hierarchy after level-wise inconsistent node removal\\
$\mathcal{H}_G$ & modified hierarchy after global inconsistent node removal\\
$\mathcal{N}_k$  & set of nodes at $k$-th level in $\mathcal{H}$\\
$\mathcal{N}$ & set of nodes (except root) in $\mathcal{H}$\\
$\mathcal{I}_L$ & set of inconsistent nodes using level-wise INF method\\
$\mathcal{I}_G$ & set of inconsistent nodes using global-INF method\\
$\mu (\mathcal{S})$ & mean of samples in set $\mathcal{S}$\\
$\sigma (\mathcal{S})$ & standard deviation of samples in set $\mathcal{S}$\\
$\mathcal{S}_k$ & set of $f^*$ values for node at $k$-th level in $\mathcal{H}$\\
$\mathcal{S}$ & set of $f^*$ values for all nodes (except root) in $\mathcal{H}$\\
$\tau_k$ & threshold limit for identifying inconsistent nodes at $k$-th level in $\mathcal{H}$\\
$\tau$ & threshold limit for identifying inconsistent nodes in $\mathcal{H}$\\
$\psi_k \geq 0$ & fitness parameter for level-wise threshold selection at $k$-th level in $\mathcal{H}$\\
$\psi \geq 0$ & fitness parameter for global threshold selection in $\mathcal{H}$\\
\hline
\end{tabular}
\end{table}

Hierarchy modification using a supervised learning approaches are also proposed in the literature \cite{babbar2013flat,tang2006acclimatizing}, where the hierarchy is gradually modified to achieve better hierarchy for improving the classification performance. These methods have an expensive evaluation costs that needs to be performed after each modification, making it computationally infeasible for large-scale settings. Hence, we do not compare our approach to these methods. Other competitive methods that involves restructuring the hierarchy are developed by us and appear in an arXiv publication \cite{naik2016taxmod}.

\section{Methods}
Table \ref{table:notDes} summarizes the common notations used in this paper. We use {\bf{bold}} letters to indicate vector variables.

\subsection{Problem Setup}
Given, a hierarchy $\mathcal{H}$ we train a binary one-vs-rest classifiers for each of the node $n\in\mathcal{N}$ --- to discriminate its positive examples from the examples of other nodes ($i.e.$, negative examples) in the hierarchy. We followed the `inclusive policy' for training classifiers, where all the descendant categories of node $n$ (including node itself) is considered as positive examples and the remaining categories as negative examples \cite{eisner2005improving}. In this paper, we have used logistic regression (LR) \cite{naik2013classifying} as the underlying base model for training. The LR objective uses logistic loss to minimize the empirical risk and $l2$-norm term (denoted by $||\cdot||_{2}^{2}$) to control the model complexity and prevent from overfitting. The objective function for training a model corresponding to node $n$ is provided in eq. (\ref{ARLR}).
\begin{equation}
\scriptsize
\min_{{\bf{\Theta_{n}}}}\Bigg[C\sum_{i=1}^{N}\log\left(1+\exp\left(-y_{i}^{n}{\bf{\Theta_{n}}}^{T}\mathbf{x}_i\right)\right)+\frac{1}{2}\left\Vert {\bf{\Theta_{n}}}\right\Vert _{2}^{2}\Bigg]\label{ARLR}
\end{equation}

For each node $n$, we solve eq. (\ref{ARLR}) to obtain the optimal weight vector denoted by ${\bf{\Theta}}_{n}$. The complete set of parameters for all the nodes $\left\{ {\bf{\Theta}}_{n}\right\} _{n\in\mathcal{N}}$ constitutes the learned model for the hierarchical top-down classifier. For LR models the conditional probability for $\hat{y}_i^{n}\in\pm1$ given its feature vector $\textbf{x}_i$ and the weight vector ${\bf {\Theta}}_n$ is given by eq. (\ref{APD}) and the classification decision function using eq. (\ref{eq:ADEC}).
\begin{equation}
\scriptsize
P(\hat{y}_i^{n}\mid{{\bf{x}}_i},{\bf{\Theta}}_n)=\dfrac{1}{\left(1+\exp\left(-y_{i}^{n}{\bf{\Theta}}_{n}^{T}{\bf{x}}_i\right)\right)}\label{APD}
\end{equation}
\begin{gather}
\scriptsize
\hat{y}_i^{n}=\begin{cases}
+1 & f_n({\bf{x}}_i) = {\bf{\Theta}}_{n}^{T}{\bf{x}}_i\ge0\\
-1 & \text{otherwise}
\end{cases}\label{eq:ADEC}
\end{gather}

For a test example with feature vector ${\bf{x}}_i$, the top-down classifier predicts the class label $\hat{y}_i\in\mathcal{L}$ as shown in eq. (\ref{eq:TDTest}), where $\mathcal{C}\left(p\right)$ denotes the set of children of node $p$. Essentially, the algorithm starts at the root node and recursively selects the best child nodes till it reaches a terminal node belonging to the set of leaf categories $\mathcal{L}$.
\begin{align}
\scriptsize
\hat{y}_i=\left\{ \begin{alignedat}{1} & \mathbf{initialize}\quad p:=root\\
 & \mathbf{while}\ p\notin\mathcal{L}\\
 & \quad p:=\mathbf{argmax}_{q\in\mathcal{C}(p)}\ f_{q}({\bf{x}}_i)\\
 & \mathbf{return}\ p
\end{alignedat}
\right\} \label{eq:TDTest}
\end{align}

\subsection{Inconsistent Node Flattening}
\paragraph*{{\bf{Motivation}}}
Gao et al. \cite{gao2011discriminative} showed that for any classifier
that correctly classifies $m$ random input-output pairs using a
set of $\mathcal{D}$ decision nodes, the generalization error bound with probability estimates greater than 1 - $\zeta$ is less than the expression shown in eq. (\ref{RCFC}).
\begin{equation}
\scriptsize
\frac{\delta r^{2}}{m}\Bigg[\sum_{n\in \mathcal{D}}(\frac{1}{\gamma_{n}^{2}})\log(4em)\log(4m)+\mathcal{\left|D\right|}\log(2m)-\log(\frac{2}{\zeta})\Bigg]\label{RCFC}
\end{equation}

where $\gamma_{n}$ denotes the margin at node $n \in \mathcal{D}$, $\delta$ is a constant term and $r$ is the radius of the ball containing the distribution's support.

This provides two significant strategies in designing our approach to reduce the generalization error:  (i) Increase the margin $\gamma_{n}$ for learned models at node $n \in \mathcal{N}$ in the hierarchy, or (ii) Decrease the number of decision nodes $|\mathcal{D}|$ involved in making the prediction. For achieving the optimum classification performance, we need to balance the trade-off between the margin $\gamma_n$ and the number of decision nodes $|\mathcal{D}|$. Two of the extreme cases for learning hierarchical classifiers are  flat and top-down methods. For flat classifiers, we have to make single decision (\emph{i.e.}, $|\mathcal{D}|$ = 1) but margin width $\gamma_{n}$ is presumably small due to the large number of leaf categories that needs to be distiguished, which makes it difficult to obtain large margin. For top-down hierarchical classifiers, we have to make a series of decisions from root to leaf nodes (\emph{i.e.}, $|\mathcal{D}| \geq 1$) but margin $\gamma_n$ is larger due to the fewer number of categories that needs to be distinguished at each of the decision nodes. Motivated by this trade-off, in this paper we propose a method that removes some of the inconsistent nodes in the hierarchy $\mathcal{H}$, and thereby, increasing the value of margin $\gamma_{n}$ for learned models at node $n$ in the hierarchy, while minimizing the number of decision nodes to classify an unlabeled test instances. 

In order to improve the effectiveness of classification we need to identify these inconsistent nodes and flatten them. We mark a node $n$ within the hierarchy as inconsistent if the value of the objective function $f_{n}^{*}$ becomes greater than a chosen threshold value. To get a more reliable estimate of the $f_{n}^{*}$, we first train the regularized LR models on a training set locally for each node and then compute the objective function on a separated validation set, which is different from the training set. The objective value on validation set for node $n$ is denoted by $f_{n}^{*}$.  We develop the following approaches for setting the threshold for flattening.

{{\bf{Level-wise Inconsistent Node Flattening}}}: In this approach, referred as Level-INF, we select a different threshold $\tau_{k}$ locally for each level $k$ in the hierarchy. Algorithm \ref{LINR} presents the level-wise approach that selects inconsistent nodes at each level in a top-down manner. The threshold $\tau_{k}$ for level $k$ is computed as the sum of mean and $\psi_k$ times the standard deviation of the set of values $\left\{ f_{n}^{*}\right\} _{n\in\mathcal{N}_{k}}$, where $\psi_k$ is a fitness parameter at level $k$ that is empirically estimated for each dataset based on $\left\{ f_{n}^{*}\right\} _{n\in\mathcal{N}_{k}}$ values (see Section \ref{empiricalStudy}) and $\mathcal{N}_{k}$ represents the set of nodes in level $k$. All nodes $n\in\mathcal{N}_{k}$ that satisfy $f_{n}^{*}>\tau_{k}$ are marked as inconsistent and added to the set of inconsistent nodes denoted by $\mathcal{I}_{L}$. This procedure is repeated for all levels of the hierarchy. Finally, we flatten the nodes in set $\mathcal{I}_{L}$ --- remove $n\in\mathcal{I}_{L}$ and corresponding edges, and add edges from children of $n$ to $n$'s parent node. The modified hierarchy thus obtained is denoted by $\mathcal{H}_{L}$. Using the modified hierarchy, we re-train a top-down classifier.

\begin{algorithm}[t!]
\scriptsize
\SetAlgoLined
 \KwData{Original Hierarchy $\mathcal{H}$, input-output ($x_i, y_i$)}
 \KwResult{Modified Hierarchy $\mathcal{H}_L$}
 Train $l2$-regularized LR model in a top-down order

/* \textbf{Set of inconsistent node, initially empty} */

 $\mathcal{I}_L$ := $\Phi$;

 \For{k := 1 $\ldots$ end$\_$level}{
  /* \textbf{Set of all nodes $f^*$ values in the level} */
  
  ${\mathcal{S}_k}$ := $\Phi$;

  \For{n $\in$ $\mathcal{N}_k$}{
  ${\mathcal{S}_k}$ := ${\mathcal{S}_k}$ $\cup$ \{$f^*_n$\}\;
  }

  $\tau_k$ := $\mu ({\mathcal{S}_k})$ + $\psi_k$$\sigma ({\mathcal{S}_k})$;

  /* \textbf{Identify inconsistent node in level} */

  \For{n $\in$ $\mathcal{N}_k$}{
      \If{$(f^{*}_{n} > \tau_k \; \& \; n \notin \mathcal{L})$}{
        $\mathcal{I}_L$ := $\mathcal{I}_L$ $\cup$ \{$n$\}\;
   }
   }
 }
/* {\textbf{New hierarchy with inconsistent node(s) removed}} */

$\mathcal{H}_L$ = $\mathcal{H}$ - \{$\mathcal{I}_L$\};

\Return{$\mathcal{H}_L$}
 \caption{Level-wise Inconsistent Node Removal}
 \label{LINR}
\end{algorithm}

{{\bf{Global Inconsistent Node Flattening}}}: Different from Level-INF approach, which sets different thresholds for each level, the global method shown in Algorithm \ref{GINR} computes a single threshold value for all levels. The threshold $\tau$ is computed as the sum of mean and $\psi$ times the standard deviation of the set of value $\left\{ f_{n}^{*}\right\} _{n\in\mathcal{N}}$, where $\psi$ is a fitness parameter that is empirically estimated for dataset considering all $\mathcal{N}$ nodes $f_n^*$ values. $\tau$ is used to identify the set of inconsistent nodes $\mathcal{I}_{G}$ in the hierarchy (\emph{i.e.}, all nodes $n$ with $f_{n}^{*}>\tau$). The hierarchy obtained by flattening the nodes present in $\mathcal{I}_{G}$ is denoted by $\mathcal{H}_{G}$. Using the modified hierarchy, we re-train a top-down classifier. In this paper we refer to this approach as Global-INF. 

\section{Experimental Evaluations}
\subsection{Datasets}
We have used text and image datasets for evaluating the performance of our proposed approaches. Various statistics of the datasets used in our experiments are listed in Table \ref{table:finaltabledataset}. All these datasets are single-labeled and the examples are mandatorily assigned to the leaf nodes in the hierarchy (although our proposed approaches is trivially extendable to datasets with multi-label and non-mandatory leaf node label assignments). For all text datasets, we have applied the tf-idf transformation with $l2$-norm normalization to the word-frequency feature vector. Description of the used data is as follows:

{\textbf{Image Datasets}}
\paragraph*{CLEF \cite{dimitrovski2011hierarchical}} Medical images annotated with medical applications codes. Each image is represented by the 80 features that are extracted using local distribution of edges.
\paragraph*{DIATOMS \cite{Dimitrovski11:jrnl}} Diatom images that was created as the part of the ADIAC project. Features for each image is created using various feature extraction techniques mentioned in \cite{Dimitrovski11:jrnl}. Further, we have preprocessed the original dataset by removing the examples that belongs to the internal nodes.
\begin{algorithm}[t!]
\scriptsize
\SetAlgoLined
 \KwData{Original Hierarchy $\mathcal{H}$, input-output ($x_i, y_i$)}
 \KwResult{Modified Hierarchy $\mathcal{H}_G$}

 Train $l2$-regularized LR model in a top-down order

/* \textbf{Set of inconsistent node, initially empty} */

 $\mathcal{I}_G$ := $\Phi$;

 /* \textbf{Set of all nodes (except root) $f^*$ values in $\mathcal{H}$} */

  ${\mathcal{S}}$ := $\Phi$;

  \For{n $\in$ $\mathcal{N}$}{
  ${\mathcal{S}}$ := ${\mathcal{S}}$ $\cup$ \{$f^*_n$\}\;
  }

  $\tau$ := $\mu ({\mathcal{S}})$ + $\psi$$\sigma ({\mathcal{S}})$;

 /* \textbf{Identify inconsistent node in $\mathcal{H}$} */

 \For{n $\in$ $\mathcal{N}$}{

      \If{$(f^{*}_{n} > \tau$ $\&$ n $\notin$ $\mathcal{L})$}{
        $\mathcal{I}_G$ := $\mathcal{I}_G$ $\cup$ \{$n$\}\;
   }
 }
/* {\textbf{New hierarchy with inconsistent node(s) removed}} */

$\mathcal{H}_G$ = $\mathcal{H}$ - \{$\mathcal{I}_G$\};

\Return{$\mathcal{H}_G$}
 \caption{Global Inconsistent Node Removal}
 \label{GINR}
\end{algorithm}

\begin{table}[t!] 
\centering 
\caption{\textbf{Dataset statistics}} 
\scriptsize
\begin{tabular}{|@{}l@{}|@{} c @{\hskip 0.050in} c @{\hskip 0.050in}  c@{\hskip 0.050in}  c @{\hskip 0.050in} c @{\hskip 0.050in}  c@{}|} 
\hline
\multicolumn{1}{|@{}c@{}|@{}}{\textbf{Dataset}} & {\textbf{\#Total Node}} & {\textbf{\#Leaf Node}} & \textbf{{Depth}} & \textbf{{\#Training}} &\textbf{ {\#Testing}} & \textbf{{\#Features}}\\
\hline
\bf{CLEF} & 88 & 63 & 4 & 10,000 & 1,006 & 80\\
\bf{DIATOMS} & 399 & 311 & 4 & 1,940 & 993 & 371\\
\bf{IPC} & 553 & 451 & 4 & 46,324 & 28,926 & 1,123,497\\
\bf{DMOZ-SMALL} & 2,388 & 1,139 & 6 & 6,323 & 1,858 & 51,033 \\
\bf{DMOZ-2010} & 17,222 & 12,294 & 6 & 128,710 & 34,880 & 381,580\\
\bf{DMOZ-2012} &  13,963 & 11,947 & 6 & 383,408 & 103,435 & 348,548\\
\hline
\end{tabular}
\label{table:finaltabledataset} 
\end{table} 
\begin{figure}[t!]
       \center
        \subfloat[][DMOZ-SMALL]{
            \includegraphics[width=.3175\linewidth,height=2.05cm]{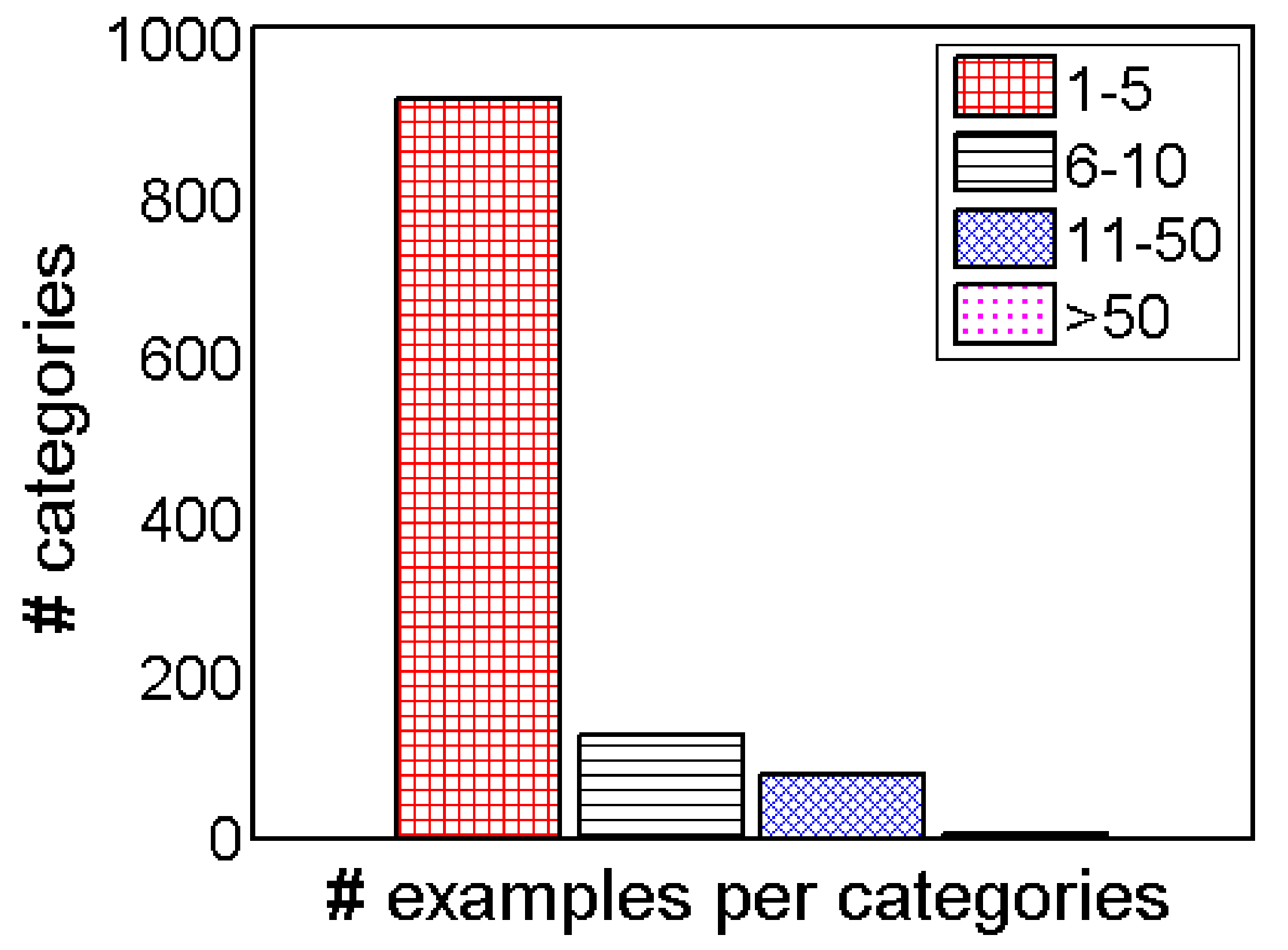}%
            \label{fig:dmoz_2010_small_dataset_plot.png}}%
        \hfil
        \subfloat[][DMOZ-2010]{
            \includegraphics[width=.3175\linewidth,height=2.05cm]{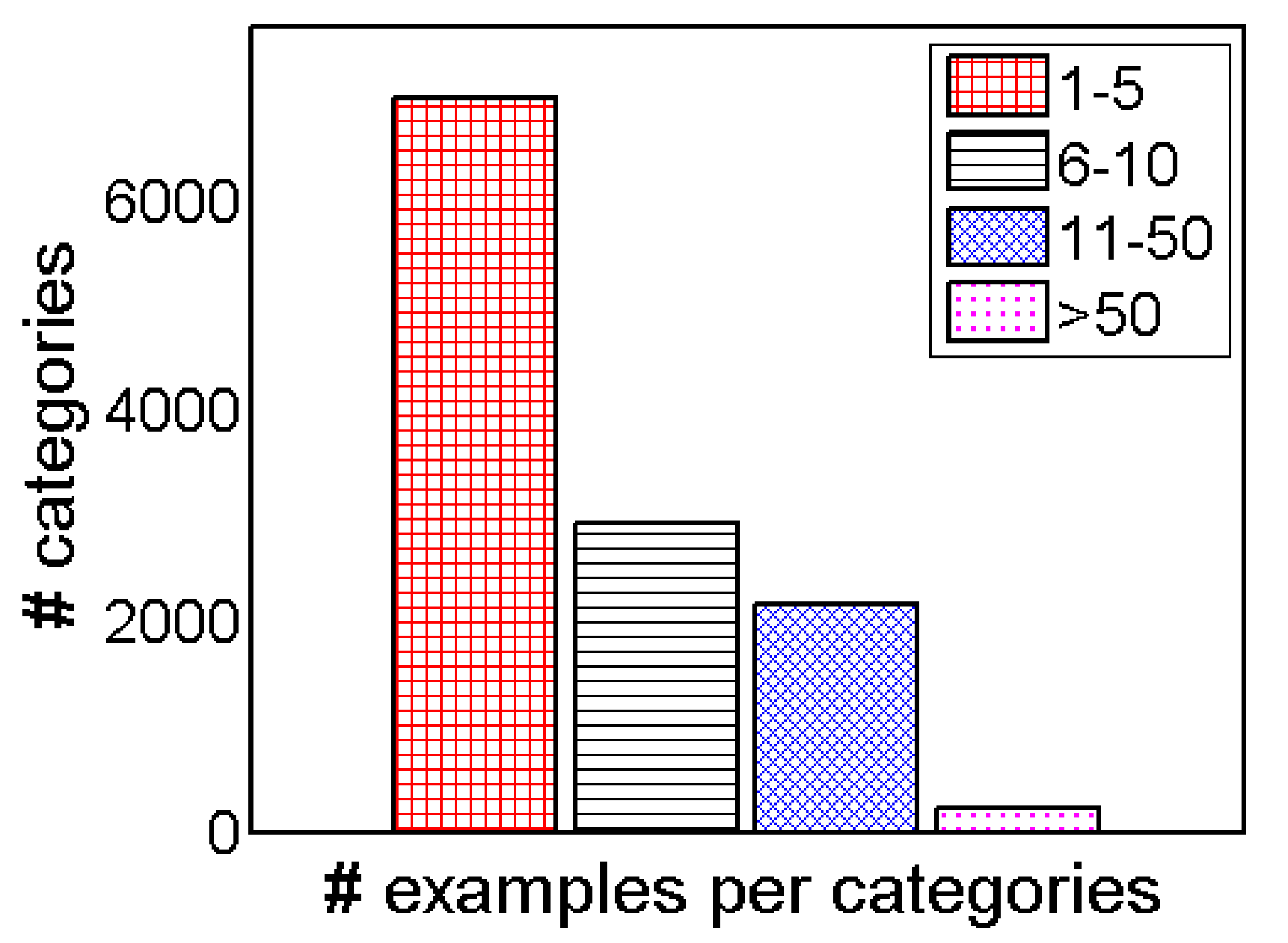}%
            \label{fig:dmoz_2010_dataset_plot.png}}%
        \hfil
        \subfloat[][DMOZ-2012]{
            \includegraphics[width=.3175\linewidth,height=2.05cm]{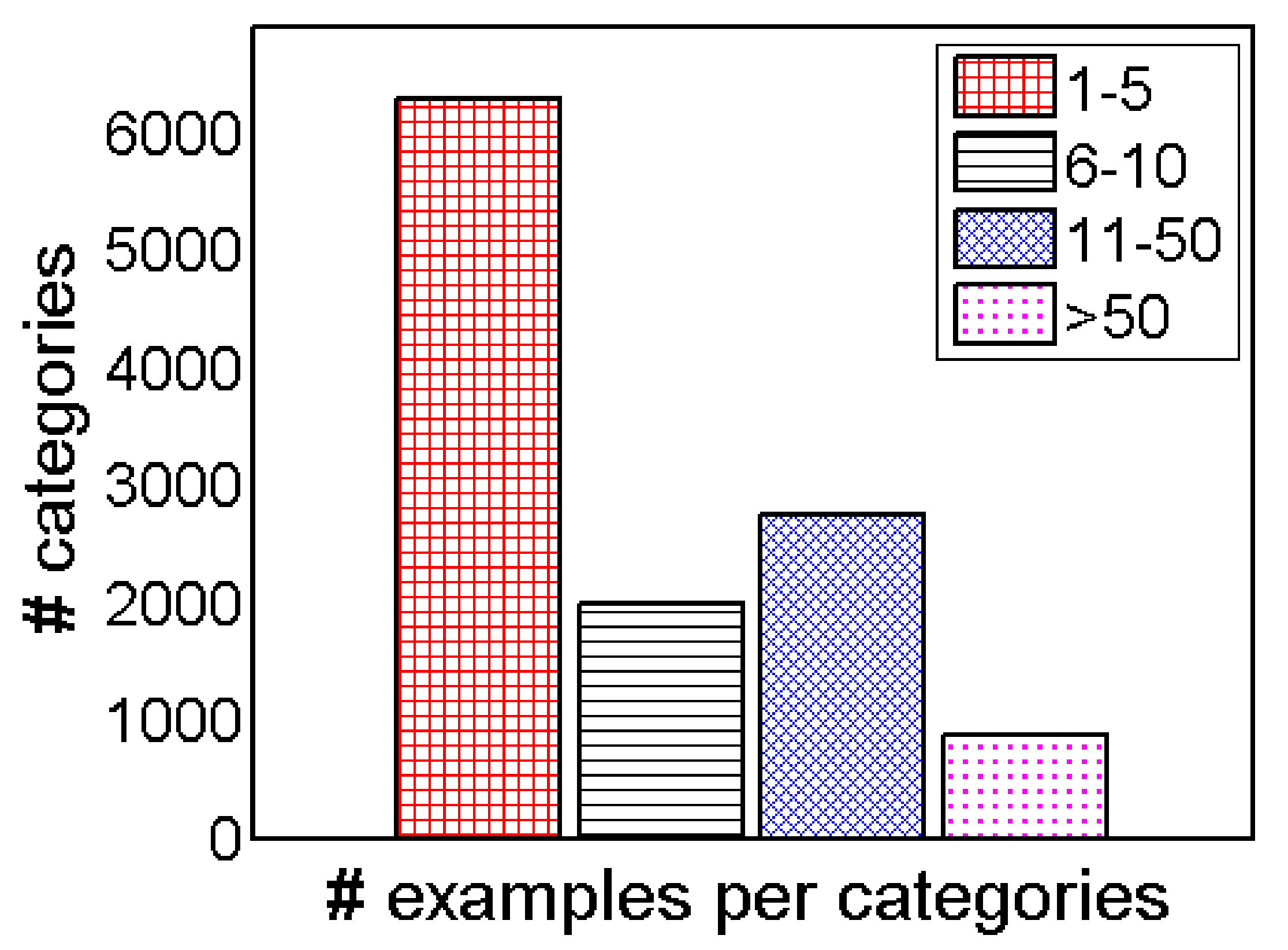}%
            \label{fig:dmoz_2012_dataset_plot.png}}%
    \caption{{{Distribution of DMOZ datasets visualizing majority of the classes with rare categories (marked in red and black).}}}
    \label{dataDistribution}
    \vspace*{-5mm}
\end{figure}
{\textbf{Text Datasets}}
\paragraph*{IPC\footnote{http://www.wipo.int/classifications/ipc/en/}} Collection of patent documents organized in international patent classification hierarchy.
\paragraph*{DMOZ-SMALL, DMOZ-2010 and DMOZ-2012\footnote{http://dmoz.org}} Multiple web documents organized into various classes using the hierarchical structure. It is subset of the web pages from open directory project and has been released as the part of the LSHTC\footnote{http://lshtc.iit.demokritos.gr/} challenge. For evaluating the DMOZ-2010 and DMOZ-2012 datasets we have used the provided test split and prediction scores are obtained using the web-portal interface\footnote{http://lshtc.iit.demokritos.gr/node/81}$^{,}$\footnote{ http://lshtc.iit.demokritos.gr/LSHTC3\_oracleUpload} that was used during the competition.

\subsection{Evaluation Metrics}
{\textbf{Flat Evaluation Measures}}:
We have used the standard metrics \cite{yang1999evaluation} micro-$F_1$ ($\mu$$F_1$) and macro-$F_1$ (M$F_1$) for evaluating the performance of various methods. To compute $\mu$$F_1$, we sum up the category specific true positives $(TP_c)$, false positives $(FP_c)$ and false negatives $(FN_c)$ for different categories and compute the score as:
\begin{gather}
\scriptsize
P = \frac{\sum_{c \in \mathcal{L}}TP_c}{\sum_{c \in \mathcal{L}}(TP_c + FP_c)}, 
R = \frac{\sum_{c \in \mathcal{L}}TP_c}{\sum_{c \in\mathcal{L}}(TP_c + FN_c)}\nonumber\\
\mu F_1 = \frac{2PR}{P + R}\end{gather}

Unlike $\mu$$F_1$, M$F_1$ gives equal weight to all the categories so that the average score is not skewed in favor of the larger categories. It is defined as follows: 
\begin{gather}
\scriptsize
P_c = \frac{TP_c}{TP_c + FP_c}, R_c = \frac{TP_c}{TP_c + FN_c} \nonumber\\
MF_1 = \frac{1}{|\mathcal{L}|}\sum_{c \in \mathcal{L}}\frac{2P_cR_c}{P_c + R_c}
\end{gather}\\
where, $|\mathcal{L}|$ is the number of leaf categories.

{\textbf{Hierarchical Evaluation Measures}}:
\label{hierEval}
Different from flat measures that penalizes each of the misclassified examples equally, hierarchical measures take into consideration hierarchical distance between the true label and predicted label for evaluating the classifier performance. In general, misclassifications that are closer to the actual class are less severe than misclassifications that are farther from the true class with respect to the hierarchy. The hierarchy based measures includes hierarchical $F_1$ ($hF_1$) (harmonic mean of hierarchical precision ($hP$), hierarchical recall ($hR$)) and tree-induced error ($TE$) \cite{dekel2004large}, which are defined as follows: 
\begin{gather}
\scriptsize 
hP = \frac{\sum_{i=1}^{N}|{\mathcal{{A}}}(\hat{y_i}) \cap {\mathcal{{A}}}({y_i})|}{\sum_{i=1}^{N}|{\mathcal{{A}}}(\hat{y_i})|}, 
hR = \frac{\sum_{i=1}^{N}|{\mathcal{{A}}}(\hat{y_i}) \cap {\mathcal{{A}}}({y_i})|}{\sum_{i=1}^{N}|{\mathcal{{A}}}({y_i})|} \nonumber\\
h F_1 = \frac{2*hP*hR}{hP + hR}\\
TE = \frac{1}{N}\sum_{i=1}^{N}\delta(\hat{y_i}, y_i)\end{gather}

where, ${\mathcal{{A}}}(\hat{y_i})$  and ${\mathcal{{A}}}({y_i})$ are respectively the set of ancestors of the predicted and true labels which include the label itself, but do not include the root node. $\delta(a, b)$ gives the length of the undirected path between categories $a$ and $b$ in the graph. For $TE$ lower scores are better, whereas for all other measures higher scores are better.

Note that for consistent evaluation of hierarchical measures we have used the original hierarchy for all methods unless where noted.

\begin{table*}[t!] 
\scriptsize
\begin{centering} 
\caption{\textbf{$\mu$$ F_1$ and $M$$F_1$ performance comparison of various TD hierarchical baselines against our proposed approaches}}
\label{table:HPC} 
\begin{tabular}{|@{\hskip 0.050in} l@{\hskip 0.050in} |r |c c c c c |c c |} 
\hline
\multicolumn{1}{|c|}{\textbf{Dataset}} & {} & \multicolumn{5}{c|}{\textbf{Hierarchical Baselines}}& \multicolumn{2}{c|}{\textbf{Proposed Approaches}}\\
\cline{3-9}
{} & {} & \textbf{TD-LR} & \textbf{TLF} & \textbf{BLF} & \textbf{MLF} & \textbf{MTA}& \textbf{Level-INF} & \textbf{Global-INF}\\
\hline
\multirow{2}{*}{\textbf{CLEF}} & \textbf{   $\mu$$F_1 (\uparrow)$} & 72.74 (0.43)  & 75.84 (0.32) & 73.76 (0.32) & X & 74.48 (0.42) & 75.25 (0.55) & \bf{77.14${^\dagger}$ (0.01)} \\
&\textbf{$M$$F_1 (\uparrow)$} & 35.92 (0.01) & 38.45 (0.65) & 40.93 (0.19) & X & 39.53 (0.74) & 39.89 (0.24) &\bf{46.54${^\ddagger}$ (0.01)} \\
\multirow{2}{*}{\textbf{DIATOMS}} & \textbf{$\mu$$F_1 (\uparrow)$} & 53.27 (0.32) & 56.93 (0.28) & 53.27 (0.24) & X & 58.36 (0.64) & 58.32 (0.64) & \bf{61.31$^{\ddagger}$ (0.53)} \\
&\textbf{$M$$F_1 (\uparrow)$} & 44.46 (0.24) & 45.17 (0.62) & 44.30 (0.64) & X & 45.21 (0.65) & 48.77 (0.12) & \bf{51.85${^\ddagger}$ (0.23)}\\
\multirow{2}{*}{\textbf{IPC}} & \textbf{$\mu$$F_1 (\uparrow)$} & 49.32 (0.32) & 51.28 (0.61) & 50.36 (0.64) & X & 51.36 (0.32) & 50.40 (0.32) &  \bf{52.30${^\dagger}$ (0.12)}\\
&\textbf{$M$$F_1 (\uparrow)$} & 42.51 (0.94) & 44.99 (0.43) & 43.74 (0.81) & X & 42.80 (0.94) & 43.26 (0.43) &  \bf{45.65${^\dagger}$ (0.11)}\\
\multirow{2}{*}{\textbf{DMOZ-SMALL}} & \textbf{$\mu$$F_1 (\uparrow)$} & 45.10 (0.23) & 45.48 (0.19) & 44.34 (0.32) & 45.80 (0.64) & 46.01 (0.74) & 45.43 (0.21) & \bf{46.61${^\dagger}$ (0.28)}\\
&\textbf{$M$$F_1 (\uparrow)$} & 30.65 (0.43) & 30.60 (0.54) & 30.94 (0.53) & 30.62 (0.32) & 30.82 (0.63) & 30.34 (0.12) & \bf{31.86${^\ddagger}$ (0.64)} \\
\multirow{2}{*}{\textbf{DMOZ-2010}} & \textbf{$\mu$$F_1 (\uparrow)$} & 40.22 (0.55) & 41.32 (0.32) & 40.34 (0.24) & 41.77 (0.56) & 41.82 (0.42) & 40.71 (0.83) & \bf{42.37} \bf{ (0.27)}\\
&\textbf{$M$$F_1 (\uparrow)$} & 28.37 (0.46) & 29.05 (0.84) & 28.41 (0.57) & 29.11 (0.13) & 29.18 (0.54) & 28.66 (0.53) & \bf{30.41} \bf{ (0.64)}\\
\multirow{2}{*}{\textbf{DMOZ-2012}} & \textbf{$\mu$$F_1 (\uparrow)$} & 50.13 (0.28) & 50.32 (0.42) & 50.11 (0.32) & 48.05 (0.39) & 50.31 (0.48) & 49.90 (0.92) & \bf{50.64} \bf{ (0.22)}\\
&\textbf{$M$$F_1 (\uparrow)$} & 29.89 (0.23) & 29.89 (0.23) & 29.73 (0.14) & 27.65 (0.48) & 30.04 (0.57) & 30.52 (0.74) & \bf{30.58} \bf{ (0.28)}\\
\hline
\end{tabular}
\par\end{centering}
    \begin{tablenotes}
      \small
      \item Table shows mean and (standard deviation) in bracket across five runs. 'X' denotes MLF not possible. The significance-test results are denoted as ${^\dagger}$ for a p-value less than 0.05 and ${^\ddagger}$ for p-value less than 0.01. We have used sign-test and wilcoxon rank test for statistical evaluation of $\mu F_1$ and $MF_1$ scores, respectively. Tests are between our best proposed approach, Global-INF and best baseline approach, MTA for single run. These statistical tests are not performed on DMOZ-2010 and DMOZ-2012 datasets because we do not have access to true labels from the online evaluation system.
    \end{tablenotes}
\end{table*}

\begin{table*}[t!] 
\scriptsize
\begin{centering} 
\caption{\textbf{Hierarchical performance comparison of various TD hierarchical baselines against our proposed approaches over original (experts defined) and modified hierarchy}}
\label{HM}
\begin{tabular}{| l |c |c c c c c |c c |} 
\hline
\multicolumn{1}{|c|}{\textbf{Dataset}} & {\textbf{Hierarchy}} & \multicolumn{5}{c|}{{\textbf{Hierarchical Baselines}}}& \multicolumn{2}{c|}{\textbf{Proposed Approaches}}\\
\cline{3-9}
{} & { } & \textbf{TD-LR} & \textbf{TLF} & \textbf{BLF} & \textbf{MLF} & \textbf{MTA}& \textbf{Level-INF} & \textbf{Global-INF}\\
\hline
&&&\multicolumn{3}{c}{{ $hF_1$ {\bf{score}}} $(\uparrow)$}&&&\\
\cline{3-9}
\multirow{2}{*}{\textbf{CLEF}} & Original & 74.52 (0.01)  & 78.24 (0.75) & 75.13 (0.46) & X & 76.01 (0.74) & 76.81 (0.59) & \bf{79.06 (0.01)} \\
& Modified & - & 77.78 (0.65) & 78.08 (0.13) & X & 77.50 (0.23) & 78.28 (0.24) &\bf{80.87 (0.13)} \\
\multirow{2}{*}{\textbf{DIATOMS}} & Original & 56.15 (0.21)  & 62.53 (0.43) & 56.14 (0.17) & X & 59.60 (0.28) & 60.03 (0.24) & \bf{62.80 (0.04)} \\
& Modified & - & 63.38 (0.24) & 57.02 (0.62) & X & 59.70 (0.14) & 59.98 (0.28) &\bf{63.88 (0.13)} \\
\multirow{2}{*}{\textbf{IPC}} & Original & 62.57 (0.32) & 64.39 (0.38) & 63.00 (0.10) & X & 63.42 (0.54) & 63.26 (0.34) & \bf{64.73 (0.12)}\\
& Modified & - & 65.48 (0.32) & 63.24 (0.41) & X & 63.14 (0.54) & 62.52 (0.38) & \bf{66.29 (0.28)} \\ 
\multirow{2}{*}{\textbf{DMOZ-SMALL}} & Original & 63.14 (0.54) & 63.17 (0.43) & 63.26 (0.52) & 63.32 (0.64) & 63.20 (0.54) & 61.98 (0.56) & \bf{63.37 (0.44)}\\
& Modified & - & 64.32 (0.50) & 63.94 (0.38) & {63.39 (0.19)} & 63.82 (0.42) & 58.02 (0.14) & \bf{64.97 (0.75)} \\ 
\multirow{1}{*}{\textbf{DMOZ-2012}} & Original & 73.04 (0.21) & 72.70 (0.17) & 73.04 (0.28) & 70.49 (0.03) & 73.03 (0.11) & 71.41 (0.38) & \bf{73.19 (0.02)}\\
\hline
&&&\multicolumn{3}{c}{{TE {\bf{score}}} $(\downarrow)$}&&&\\
\cline{3-9}
\multirow{2}{*}{\textbf{CLEF}} & Original & 1.26 (0.01)  & 1.08 (0.08) & 1.23 (0.03) & X & 1.13 (0.09) & 1.15 (0.05) & \bf{1.04 (0.03)} \\
& Modified & - & 0.89 (0.07) & 0.88 (0.04) & X & 0.90 (0.04) & 0.94  (0.01) &\bf{0.71 (0.09)}\\
\multirow{2}{*}{\textbf{DIATOMS}} & Original & 1.76 (0.01)  & 1.49 (0.01) & 1.76 (0.03) & X & 1.60 (0.03) & 1.60 (0.06) & \bf{1.49 (0.02)} \\
& Modified & - & 1.28 (0.02) & 1.32 (0.02) & X & 1.14 (0.06) & 1.16  (0.02) &\bf{1.08 (0.08)} \\
\multirow{2}{*}{\textbf{IPC}} & Original & 2.23 (0.02) & 2.12  (0.04) & 2.20  (0.01) & X & 2.22  (0.06) & 2.19 (0.01) & \bf{2.10 (0.02)}\\
& Modified & - & 1.64 (0.01) & 1.58 (0.04) & X & 1.80 (0.06) & 1.83 (0.03) & \bf{1.38 (0.02)} \\
\multirow{2}{*}{\textbf{DMOZ-SMALL}} & Original & 3.55 (0.04) & 3.55  (0.02) & 3.53  (0.03) & 3.53  (0.06) & 3.51  (0.08) & 3.65 (0.06) & \bf{3.50  (0.02)}\\
& Modified & - & 2.96 (0.05) & 2.90 (0.01) & 2.62 (0.03) & 2.68 (0.03) & 2.82 (0.02) & \bf{2.37 (0.03)}\\
\multirow{1}{*}{\textbf{DMOZ-2010}} & Original & 3.69  (0.03) & 3.58  (0.01) & 3.68  (0.10) & 3.56  (0.08) & 3.61  (0.02) & 3.74 (0.04) & \bf{3.53  (0.01)}\\
\hline
\end{tabular}
\label{table:HPCte} 
\par\end{centering}
    \begin{tablenotes}
      \small
      \item Table shows mean and (standard deviation) in bracket across five runs. 'X' denotes MLF not possible. Evaluations for DMOZ-2010 and DMOZ-2012 datasets cannot be performed on new hierarchy as it is not supported by the web-portal. Further, $hF_1$ for DMOZ-2010 and TE score for DMOZ-2012 dataset is not available from the online evaluation system.
    \end{tablenotes}
\end{table*}

\subsection{Experimental Protocol}
In all the experiments, we have divided the training dataset into train and small validation dataset in the ratio 90:10. Each experiment was run five times with different sets of train and validation split chosen randomly. Testing is done on an independent held-out dataset as provided by these benchmarks. The model is trained by choosing mis-classification penalty parameter ($C$) in the set [$10^{-3}$, $10^{-2}$, $10^{-1}$, $1$, $10^{1}$, $10^{2}$, $10^{3}$]. The best parameter is selected using a validation set. The best parameters are used to re-train the models on the entire training set and the performance is measured on a held-out test set.  For the INF methods,  we compute and save the $f^*_n$ value for each node in the hierarchy using a validation set. Setting the threshold as $\mu + \psi\sigma$ (or $\mu + \psi_k\sigma$ for $k$-th level in Level-INF approach), we remove the inconsistent nodes where best value of fitness parameter $\psi$ (or $\psi_k$) is computed empirically for each dataset (see Section \ref{empiricalStudy}). All  experiments were conducted using a modified version of liblinear\footnote{http://www.csie.ntu.edu.tw/$\sim$cjlin/liblinear/} software \cite{REF08a} and were run on ARGO, a research computing cluster provided by the Office of Research Computing (URL: http://orc.gmu.edu), at George Mason University, VA.

\section{Comparative Approaches}
\subsection{Flat Methods}
\paragraph*{{\bf{Logistic Regression (LR)}}} We train binary one-versus-rest regularized LR classifiers for each of the leaf categories, ignoring the hierarchical structure. The prediction decision $\hat{y}$ for unlabeled test instance {\textbf{x}} is based on the maximum prediction score achieved when compared across the one-versus-rest classifiers as shown in eq. (\ref{flatprediction}).
\begin{equation}
\hat{y} = \underset{{n \; \in \;\mathcal{L}}}{{\bf{argmax\;}}} {\bf{\Theta}}_n^T{\bf{x}}
\label{flatprediction}
\end{equation}
\paragraph*{{\bf{Error Correcting Output Codes (ECOC) \cite{ghani2000using}}}} This approach combines binary classifiers to exploit correlations and correct errors. Codewords are generated randomly with bits assigned for representing the hierarchical information between the categories. Experiments were done with codeword length varying from 32 to 1024 bits depending on the dataset. For testing an unlabeled example, the output codeword is compared to the codeword of each categories, and the one with the minimum hamming distance is selected as the class label for that example.
\subsection{Top-Down (TD) Hierarchical Methods}
For all TD hierarchical baselines we train a binary one-vs-rest classifiers for each of the node (except root) in the hierarchy and predictions are made starting from the root node and recursively selecting the best scoring child nodes until a leaf node is reached (see eq. (\ref{eq:TDTest})). Depending upon the hierarchy that we use during the training and prediction process, we compare with the following baselines.
\paragraph*{{\bf{Top-Down Logistic Regression (TD-LR)}}} Original hierarchy provided by the domain experts is used for classifiers training and label prediction.
\paragraph*{{\bf{Level flattening}}} Modified hierarchy obtained by flattening different level(s) is used instead of original hierarchy. Depending on level(s) flattened we have Top Level Flattening (TLF), Bottom Level Flattening (BLF), Multiple Level Flattening (MLF) hierarchy as discussed in Section \ref{sec:lit:hm}.
\paragraph*{{\bf{Maximum-margin based Taxonomy Adaptation (MTA)}}} Original hierarchy is modified using the margin value computed at each node in the hierarchy as described in Babbar et al. \cite{babbar2013maximum}.

\section{Results and Discussion}
\label{resDis}
\subsection{Comparison to Top-down (TD) Hierarchical Baselines}
\begin{figure}[t]
       \center
        \subfloat[][{{$\mu F_1$}}]{
            \includegraphics[width=.475\linewidth,height=2.75cm]{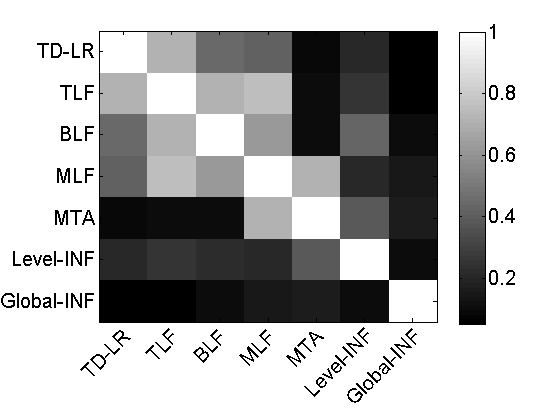}%
            \label{fig:MicroF1_heatMap}}%
        \hfil
        \subfloat[][{{$MF_1$}}]{
            \includegraphics[width=.475\linewidth,height=2.75cm]{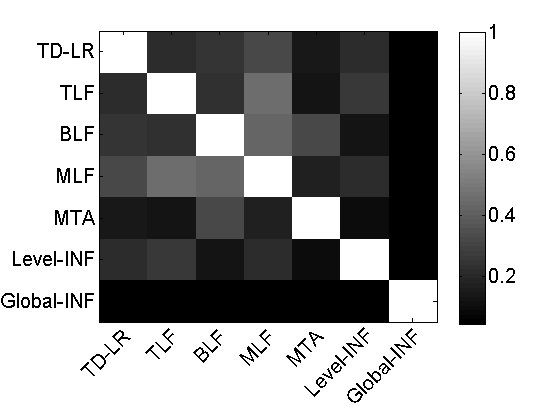}%
            \label{MacroF1_heatMap}}%
   \caption{{{P-values comparison of different approaches for DMOZ-SMALL dataset. Significant improvements are denoted by dark black color. }}}
    \label{Heatmap}
    \vspace*{-5mm}
\end{figure}

{{\bf{Performance based on Flat Metrics}}}: Table \ref{table:HPC} presents the $\mu F_1$ and $MF_1$ performance comparison 
of our proposed hierarchical modification approaches  with TD-LR (involves no hierarchy modification) and comparative TD hierarchy modification approaches as baselines. We 
see that our proposed approach Global-INF consistently outperforms all other approaches for the different datasets across all metrics. 
For the image datasets we see a relative performance improvement upto 7$\%$ in M$F_1$ on comparing Global-INF with the best TD modification baseline $i.e.$, MTA. To 
validate the performance improvement we conducted pairwise statistical significance tests between our best approach, Global-INF and best TD baseline for 
all datasets except DMOZ-2010 and DMOZ-2012, where true test labels (and class-wise performance)
are not available from the online evaluation. Specifically, we compute 
sign-test for $\mu F_1$ \cite{yang1999re} and non-parametric wilcoxon rank test for M$F_1$ scores (it should be noted that significance tests are between two approaches for single run). In Figure \ref{Heatmap} we  present the 
pairwise statistical comparisons for different 
approaches studied here on the  DMOZ-SMALL dataset. The Global-INF consistently 
outperforms other baseline approaches studied here. 

Overall, results of statistical tests shows that Global-INF approach significantly outperforms the best baseline (and hence other baselines) for all the datasets (see Table \ref{table:HPC}). 

On comparing our two proposed approaches -- Global-INF has better performance over Level-INF. This is because the Level-INF approach strictly enforces some of the nodes to be flattened from each levels although there $f^*_n$ value may be much lower than the other nodes at different levels in the hierarchy and vice-versa. In contrast, Global-INF approach takes all nodes $f^*_n$ values into consideration while making a decision and hence it determines a better set of inconsistent nodes. MTA approach has poor performance due to the similar issues as with Level-INF approach. Performance of level flattening approaches, viz., TLF, BLF and MLF, suffers because these methods remove the entire level(s) in the hierarchy and do not take into consideration whether any node in that level is important for HC. TD-LR approach has the worst performance because of the inconsistent nodes present in the original hierarchy that are negatively impacting the generalization capabilities of learned models at the higher levels (see Section \ref{LWMis}), which results in error propogation.

{{\bf{Performance based on Hierarchical Metrics}}}:
Hierarchical evaluation metrics $hF_1$ and $TE$ (described in Section \ref{hierEval}) compute errors for misclassified examples based on 
the definition of a defined hierarchy. As such,   Table \ref{HM} presents  the $hF_1$ and 
TE score for all TD approaches 
evaluated over the original hierarchy and the modified hierarchy (obtained by flattening). 
We can see that our proposed approach, Global-INF outperforms other 
approaches because
it is able to identify a better set of inconsistent nodes. 
On comparing the classification performance over the 
original hierarchy and the modified hierarchy, we can see that for most 
of the approaches classification on modified hierarchy shows  an improved  performance. This is
because flattening of hierarchies results in the reduction of
hierarchical path length for mis-classified examples contributing to performance
improvement. 


\begin{figure}[t]
\centering
    \includegraphics[width=0.5\textwidth,height=3.0cm]{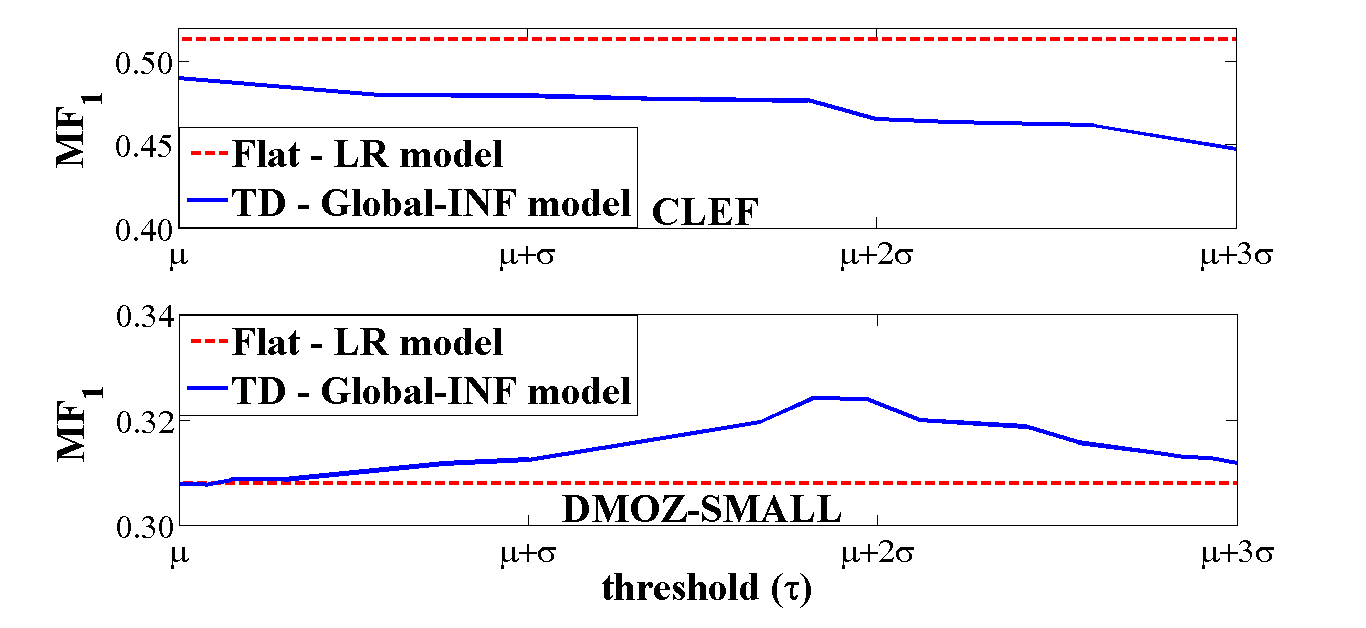}
  \caption{{$MF_1$ performance comparison of flat LR approach (marked in dotted red) against best TD approach, Global-INF (marked in solid blue) with different selection of threshold ($\tau$) for CLEF and DMOZ-SMALL datasets. Validation data is used for plotting the graph.}}
  \label{VINR}
    \vspace*{-5mm}
\end{figure}

\begin{figure}[t]
       \center
        \subfloat[][CLEF]{
            \includegraphics[width=.475\linewidth,height=2.75cm]{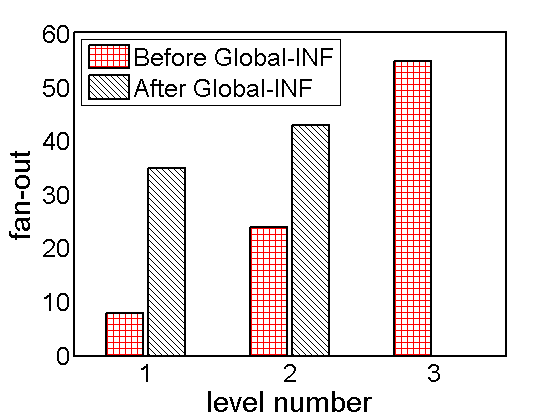}%
            \label{clef_fanout}}%
        \hfil
        \subfloat[][DMOZ-SMALL]{
            \includegraphics[width=.475\linewidth,height=2.75cm]{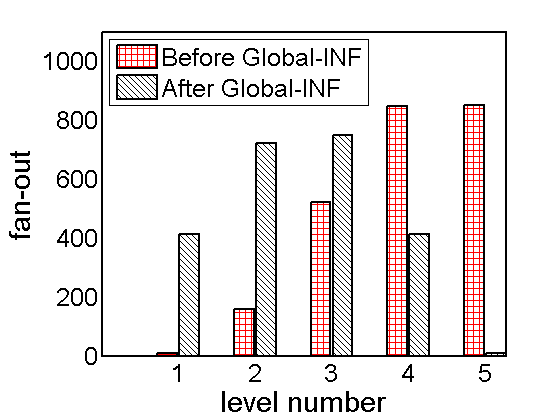}%
            \label{dmoz_small_fanout}}%
   \caption{{{Total fan-out ($\#$ children) at each level for CLEF and DMOZ-SMALL datasets before and after flattening inconsistent nodes using Global-INF approach.}}}
    \label{table:fanout}
    \vspace*{-5mm}
\end{figure}

\subsection{Empirical Study for Threshold $(\tau)$ Selection}
\label{empiricalStudy}
Figure \ref{VINR} shows the $M F_1$ performance comparison of flat LR approach against our best TD approach, Global-INF with varying selection of threshold $(\tau)$ in the interval [$\mu$, $\mu + 3\sigma$] (performance deteriorates after $\mu + 3\sigma$) with step-size 0.1$\sigma$ for CLEF and DMOZ-SMALL datasets. We choose these datasets for evaluation because they have different data characteristics. The CLEF dataset is well balanced and does not suffer from the rare categories issue, whereas DMOZ-SMALL dataset is highly imbalanced and majority of the classes belong to rare categories (\emph{i.e.}, having $\leq$ 10 examples) as shown in Figure \ref{dataDistribution}(a). In order to identify the set of inconsistent nodes in the hierarchy, we compare the computed $f_n^*$ value of each internal node with the chosen threshold $(\tau)$  and mark the node as inconsistent iff $f_n^* > \tau$. It can be seen from the figure that for CLEF dataset, performance improves as the threshold ($\tau$) decreases giving intuition that $\tau$ should be kept smaller $i.e.,$ removing more internal nodes from the hierarchy (enforcing flat structure) is better and hence reducing the threshold value $\tau$ can possibly lead to better results. However, for the DMOZ-SMALL dataset, performance first increases and than decreases with maximum performance achieved at $\tau = \mu + 1.8\sigma$. This behavior suggests that for imbalanced data distribution with potentially large number of rare categories, we should generally keep the threshold higher. It helps to leverage the hierarchical information while reducing error propagation by removing inconsistent nodes. The best threshold for a specific dataset can be chosen empirically using a small validation set as done in this study. 

To further understand the behavior of modified hierarchy using Global-INF approach, we analyzed the datasets in terms of level-wise fan-out ($\#$ children) in the hierarchy, before and after removing inconsistent nodes. We can see from the Figure \ref{table:fanout} that with both datasets maximum flattening take place at higher levels in the hierarchy, which results in increased fan-out value. Reason for inconsistencies at higher levels is the presence of many dissimilar classes beneath each node, which makes it comparatively difficult to learn generalized classifiers resulting in higher $f_n^*$ values (i.e., inconsistent nodes marked for removal). 

\begin{table}[t!] 
\begin{centering}
\caption{\textbf{Level-wise error for TD-LR and Global-INF approach}}
\label{table:LP} 
\scriptsize
\begin{tabular}{|@{} l @{}| @{}c@{}| c c@{\hskip 0.050in}| c c@{\hskip 0.050in}|} 
\hline
 \multicolumn{1}{|c|@{}}{\textbf{Dataset}}& {\textbf{Level}}  & \multicolumn{2}{c|}{\textbf{TD-LR}} & \multicolumn{2}{c|}{\textbf{Global-INF}}\\
\cline{3-6}
 & {\textbf{no.}} & \textbf{error $(\downarrow)$} &  \textbf{$\#$ ME} & \textbf{error $(\downarrow)$} &  \textbf{$\#$ ME}\\
\hline
\multirow{3}{*}{\textbf{CLEF}} & {L-1} & 21.27 (0.63) & 214 & 20.18 (0.26) & 203\\
 & {L-2}  & 07.71 (0.42) & 240 & 07.34 (0.29) & 224 \\
 & {L-3}  & 11.30 (0.16) & 274 & 05.66 (0.13)  & 227 \\
\hline 
\multirow{5}{*}{\textbf{DMOZ-SMALL}} & {L-1} & 42.47 (0.32) & 789 & 39.83 (0.17)  & 740\\
 & {L-2}  & 14.45 (0.62) & 921 & 12.91 (0.21) & 855\\
 & {L-3}  & 15.14 (0.34) & 972 & 17.99 (0.14) & 968\\
 & {L-4}  & 12.32 (0.02)  & 1001 & 07.57 (0.10)  & 991\\
 & {L-5}  & 15.66 (0.05) & 1020 & 33.33 (0.04) & 992\\
\hline
\end{tabular}
\par\end{centering}
    \begin{tablenotes}
      \small
      \item Table shows mean and (standard deviation) of error rate across five runs. $\#$ ME denotes the average number of misclassified examples upto that level.
    \end{tablenotes}
\end{table}

\subsection{{{Level-wise Misclassification Error}}}
\label{LWMis}
Table \ref{table:LP} shows the level-wise error analysis that is obtained for TD-LR and our best approach, Global-INF for CLEF and DMOZ-SMALL datasets. We can see that at higher levels, the Global-INF approach misclassifies fewer examples (shown in \#ME column) that results in less error propagation down the levels, and hence better overall performance. This experiment supports our hypothesis that Global-INF approach identifies better set of inconsistent nodes that helps in minimizing the error propagation. Results for other TD baselines are not shown in the paper for brevity.

\begin{table}[t]
\begin{centering} 
\caption{\textbf{$MF_1$ and $hF_1$ performance comparison between Global-INF and flat baselines with varying distribution of training examples per class}}
\label{table:hierarchyUseful} 
\scriptsize
\begin{tabular}{|@{}l@{}|@{}c@{}| @{\hskip 0.05cm}c@{\hskip 0.05cm} c @{\hskip 0.05cm}|@{\hskip 0.05cm}c@{\hskip 0.05cm} c@{\hskip 0.05cm}|@{\hskip 0.05cm} c@{\hskip 0.05cm} c @{\hskip 0.05cm}|} 
\hline
\multicolumn{1}{|@{}l@{}|}{}&\multicolumn{1}{@{}c@{}|@{\hskip 0.05cm}}{\textbf{\# Train}} & \multicolumn{2}{c@{\hskip 0.05cm}|@{\hskip 0.05cm}}{\textbf{Best Proposed}} & \multicolumn{4}{c@{\hskip 0.05cm}|}{\textbf{Flat Baselines}} \\
\cline{3-8}
\multicolumn{1}{|@{}c@{}|}{\textbf{Dataset}}&\multicolumn{1}{@{}c@{}|@{\hskip 0.05cm}}{\textbf{example}} & \multicolumn{2}{c@{\hskip 0.05cm}|@{\hskip 0.05cm}}{\textbf{Global-INF}} & \multicolumn{2}{c@{\hskip 0.05cm}|@{\hskip 0.05cm}}{\textbf{LR}} & \multicolumn{2}{c@{\hskip 0.05cm}|}{\textbf{ECOC}}\\
\cline{3-8}
\multicolumn{1}{|@{}l@{}|}{}& \multicolumn{1}{@{}c@{}|@{\hskip 0.05cm}}{\textbf{per class}} & \multicolumn{1}{c@{\hskip 0.05cm}}{$MF_1$} & \multicolumn{1}{c@{\hskip 0.05cm}|@{\hskip 0.05cm}}{$hF_1$} & \multicolumn{1}{c@{\hskip 0.05cm}}{$MF_1$} & \multicolumn{1}{c@{\hskip 0.05cm}|@{\hskip 0.05cm}}{$hF_1$} & \multicolumn{1}{c@{\hskip 0.05cm}}{$MF_1$} & \multicolumn{1}{c@{\hskip 0.05cm}|}{$hF_1$}\\
\hline
\multirow{4}{*}{\textbf{CLEF}} & 6-10 & {\bf{36.12}}& {\bf{66.45}} & 35.14 & 62.34 & 36.16 & 63.45\\
& 11-50 & 45.52 & 76.13 & {\bf{45.92}} & {\bf{77.04}} & 45.36 & 76.40\\
& $>$50 & 52.24 & 84.12 & {\bf{55.92}} & {\bf{87.24}} & 53.24 & 85.90\\
& {\bf{avg.}} & 46.99 & 79.00 & {\bf{51.31}}${^\ddagger}$ & {\bf{80.58}} & 50.02 & 79.34\\
\hline
\multirow{4}{*}{\textbf{DIATOMS}} & 6-10 & {\bf{41.08}} & {\bf{48.92}} & 38.92 & 46.32 & 39.14 & 47.44\\
& 11-50 & 42.82 & {\bf{62.72}} & {\bf{44.18}} & 62.45 & 43.82 & 59.26\\
& $>$50 & 53.17 & 64.10 & {\bf{57.24}} & {\bf{68.10}} & 52.21 & 63.28\\
& {\bf{avg.}} & 51.85 & 62.80 & {\bf{54.17}}${^\ddagger}$ & {\bf{63.50}} & 48.82 & 61.92\\
\hline
\multirow{3}{*}{\textbf{IPC}} & 11-50 & 43.64 & {\bf{62.45}} & {\bf{43.94}} & 60.92 & 42.98 & 60.01\\
& $>$50 & 47.28 & 68.70 & {\bf{49.44}} & {\bf{68.95}}& 47.21 & 67.96\\
& {\bf{avg.}} & 45.65 & {\bf{65.73}} & {\bf{45.74}} & 64.00 & 44.65 & 61.84\\
\hline
\multirow{5}{*}{\textbf{DMOZ-SMALL}} & $\leq$5 & {\bf{28.77}} & {\bf{51.86}} & 27.02 & 46.81 & 27.12 & 47.35\\
& 6-10 & {\bf{55.55}} & {\bf{67.47}} & 54.76 & 65.40 & 54.18 & 63.28\\
& 11-50 & 72.26 & 78.74 & {\bf{72.60}} & {\bf{80.12}}& 72.02 & 78.53\\
& $>$50 & 69.43 & 86.70 & {\bf{71.44}} & {\bf{88.95}} & 69.80 & 85.89\\
& {\bf{avg.}} & {\bf{31.86}}${^\dagger}$ & {\bf{63.37}} & 30.80 & 60.87 & 30.10 & 60.50\\
\hline
\multirow{5}{*}{\textbf{DMOZ-2010}} & $\leq$5 & {\bf{18.23}} & {\bf{53.59}} & 14.35 & 48.13 & 10.46 & 47.47\\
& 6-10 & {\bf{23.03}} & {\bf{55.76}} & 22.62 & 51.84 & 20.74 & 49.63\\
& 11-50 & 42.56 & {\bf{62.39}} & {\bf{43.26}} & 61.85 & 41.92 & 60.44\\
& $>$50 & 70.74 & 77.51 & {\bf{73.20}} & {\bf{81.51}} & 68.92 & 77.18\\
& {\bf{avg.}} & {\bf{28.41}}${^\dagger}$ & {\bf{56.17}} & 27.06 & 53.94 & 26.12 & 49.24\\
\hline
\multirow{5}{*}{\textbf{DMOZ-2012}} & $\leq$5 & \bf{10.28} & \bf{50.56} & 8.78 & 48.01 & 7.41 & 47.14 \\
& 6-10 & \bf{20.37} & \bf{50.71} & 18.84 & 48.82 & 18.32 & 48.26\\
& 11-50 & 37.19 & 73.16 & \bf{37.98} & \bf{73.24} & 35.72 & 72.73\\
& $>$50 & 53.20 & 79.73 & \bf{55.72} & \bf{84.92} & 50.23 & 78.10\\
& {\bf{avg.}} & {\bf{29.14}}${^\ddagger}$ & {\bf{68.24}} & 27.04 & 66.45 & 26.64 & 65.10\\
\hline
\end{tabular}
\par\end{centering}
    \begin{tablenotes}
      \small
      \item CLEF, DIATOMS and IPC datasets does not have any categories with $\leq$5 examples ($\leq$10 for IPC). The significance-test results are denoted as ${^\dagger}$ for a p-value less than 0.05 and ${^\ddagger}$ for p-value less than 0.01. Wilcoxon rank test is used for statistical evaluation of $MF_1$ scores. Tests are between Global-INF and best flat, LR approach.
   \end{tablenotes}
\end{table}

\subsection{Comparison to Flat Baselines}
Table \ref{table:hierarchyUseful} shows the $MF_1$ and $hF_1$ performance comparison of our best approach, Global-INF against flat baselines -- LR and ECOC approach. For easier analysis, we have showed the results for datasets separated by varying distribution of training size (for evaluating DMOZ-2010 and DMOZ-2012 datasets we have used a separate held out dataset because we don't know the actual labels of test dataset from the online evaluation). We show the results for $MF_1$ because it gives equal importance to all the classes while evaluation and hence provides better essence of the results for datasets with skewed distribution. For computing $hF_1$, we have used the original hierarchy for consistent evaluation. As we can see from the table, the LR approach outperforms Global-INF approach for CLEF, DIATOMS and IPC datasets because these datasets are well balanced and have smaller number of categories. However, for the DMOZ datasets, our approach Global-INF has better performance because hierarchical structure provides useful information for categorizing classes with rare categories. Within the DMOZ datasets, rare categories make up more than 80$\%$ of the classes as shown in Figure \ref{dataDistribution}. The ECOC approach has the worst performance because the codewords used in our experiments are chosen randomly and merging of categories may require non­linear discriminants instead of the linear classifiers used in this paper. 

\subsection{Computational Run Time}
\begin{figure}[t]
       \center
        \subfloat[][Small Datasets]{
            \includegraphics[width=.475\linewidth,height=2.75cm]{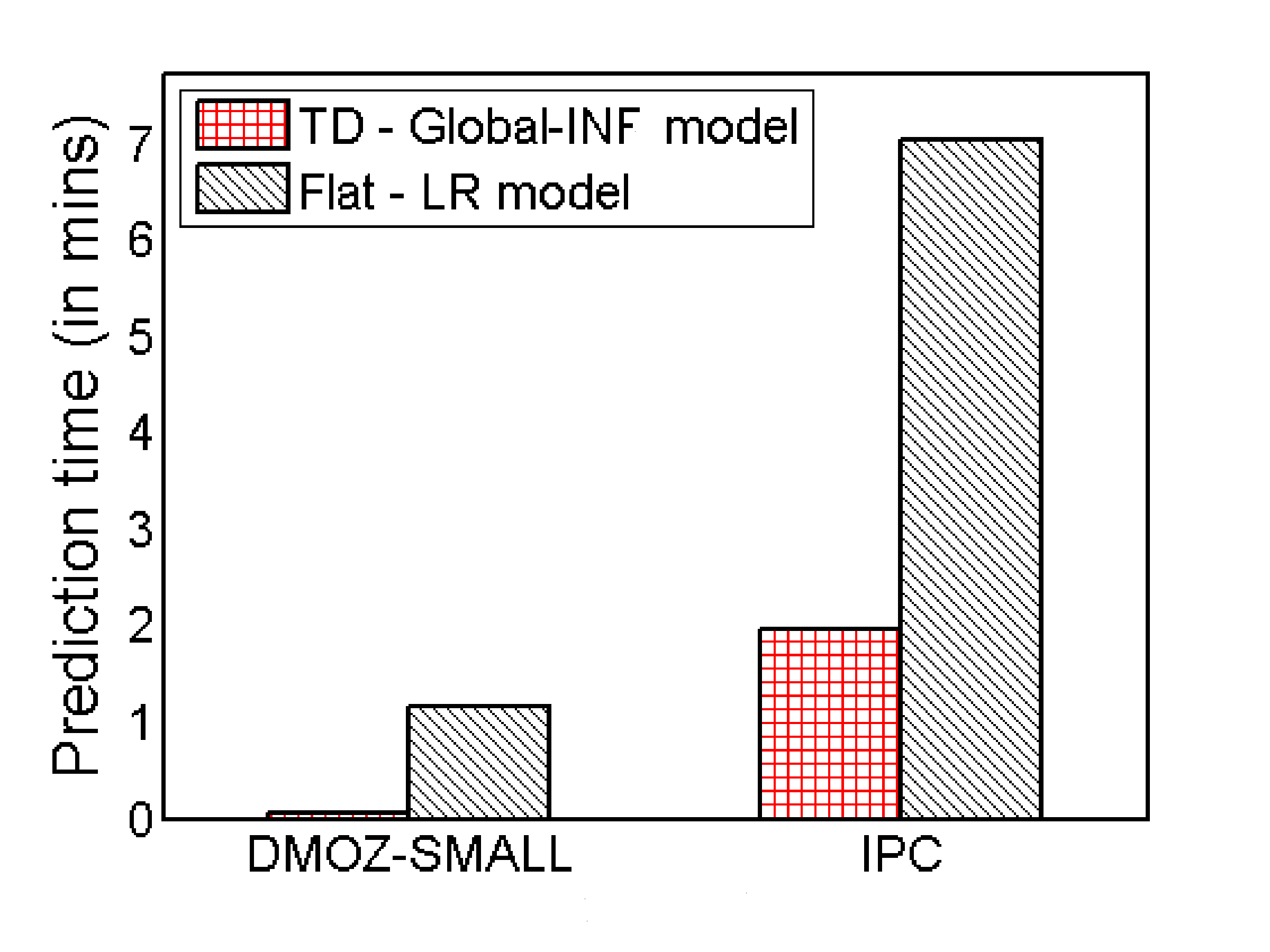}%
            \label{clef_fanout}}%
        \hfil
        \subfloat[][Large Datasets]{
            \includegraphics[width=.475\linewidth,height=2.75cm]{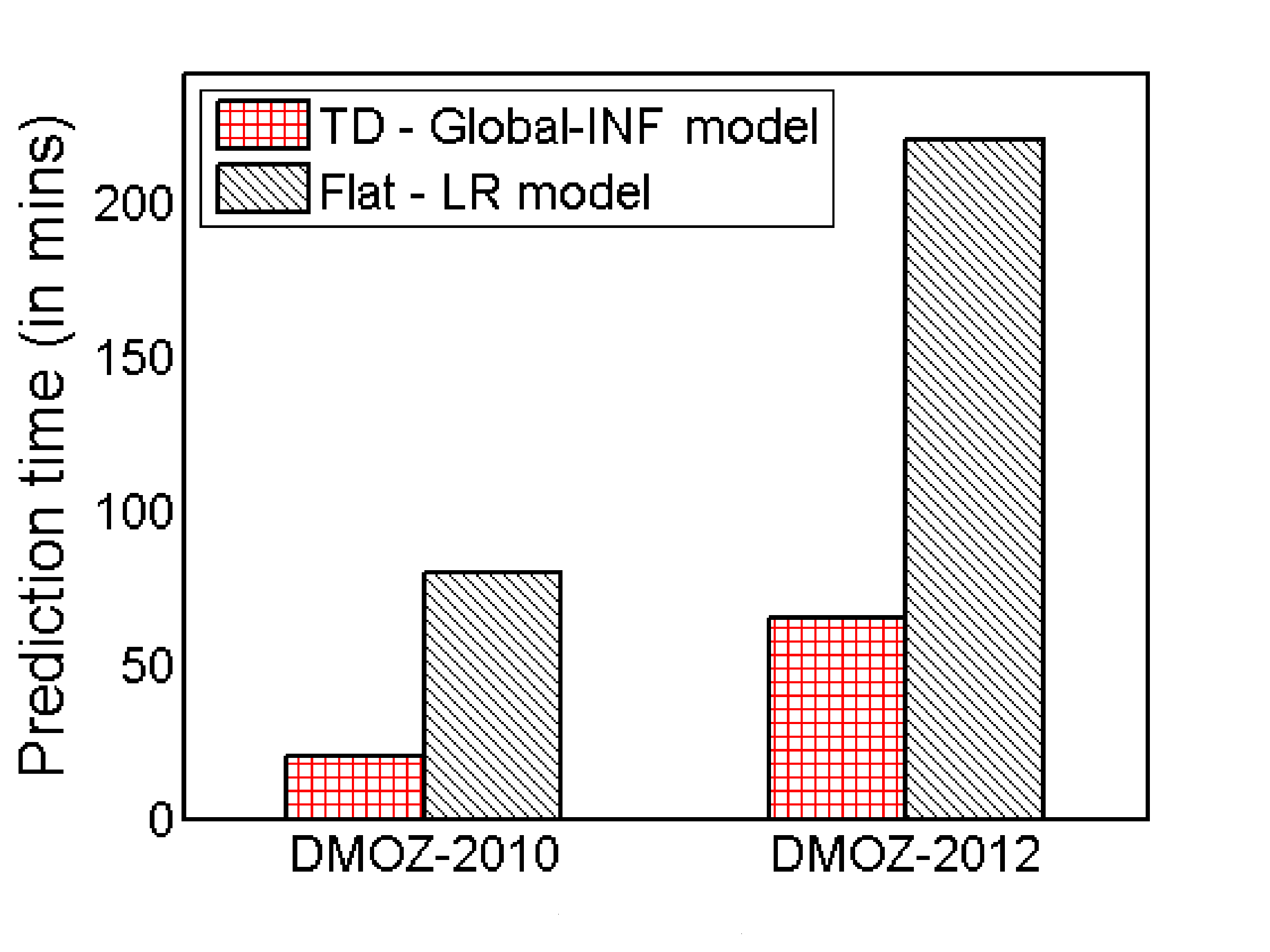}%
            \label{dmoz_small_fanout}}%
   \caption{{{Prediction runtime comparison (in mins) between Global-INF and LR approach. Image datasets have small difference hence omitted.}}}
    \label{table:predictperformance}
\end{figure}

Although, the  flat LR approach outperforms Global-INF approach for some datasets in terms of classification performance, their prediction runtime is significantly higher and it can be untenable  for large-scale problems \cite{gopal2013recursive,liu2005support}. The prediction runtime comparison of Global-INF and LR approach is shown in Figure \ref{table:predictperformance}. As expected, Global-INF approach has comparatively lower prediction runtime (upto 4x improvement). The difference is significant for large-scale datasets (DMOZ-2010 and DMOZ-2012). For completeness, we also report the total training runtime in Table \ref{table:trTime}. The Global-INF approach has higher training runtime due to the overhead involved with classifiers re-training after hierarchy modification and also involves training one-vs-rest binary classifiers for internal nodes in addition to leaf categories. Nevertheless, both flat and TD approaches are trivially parallelizable due to decoupling ($i.e.$, no interactions) between the classifiers learnt at different nodes $n$ in the hierarchy. For reporting training runtime, we trained classifiers in parallel across multiple compute nodes in the cluster and sum up the time taken at each node. In our experiments, we choose expensive one-vs-rest binary classifiers over comparably cheaper one-vs-sibling binary classifiers because our preliminary experiments showed better results with one-vs-rest approach. It should also be noted that there is no significant difference between the prediction and training runtime of different TD approaches, and hence we do not report them here. 

\begin{table}[t]
\begin{centering} 
\caption{\textbf{Total training runtime comparison (in mins) between Global-INF and LR approach}}
\label{table:trTime} 
\scriptsize
\begin{tabular}{|@{} l @{} | @{\hskip 0.05cm}c@{\hskip 0.1cm} c@{\hskip 0.1cm} c@{\hskip 0.1cm} c @{\hskip 0.1cm}c @{\hskip 0.1cm}c@{\hskip 0.05cm}|} 
\hline
\multicolumn{1}{|@{}c@{}|}{\textbf{}}& {\textbf{CLEF}} &  {\textbf{DIATOMS}} & {\textbf{IPC}} & {\textbf{DMOZ-SMALL}} & {\textbf{DMOZ-2010}} & {\textbf{DMOZ-2012}}\\
\hline
\multicolumn{1}{|@{}c@{}|}{\textbf{Global-INF}}& 3 & 10 & 830 & 68 & 25,462 & 63,000\\
\multicolumn{1}{|@{}c@{}|}{\textbf{LR}}& 1 & 3 & 658 & 46 & 15,248 & 46,124\\
\hline
\end{tabular}
\par\end{centering}
\end{table}

\section{Conclusion and Future Work}
In this paper, we proposed two different approaches for hierarchy modification that restructures the hierarchy by flattening most prominent set of inconsistent nodes, thereby improving the hierarchy representation which is more suited for HC. Performance evaluation on wide range of datasets over the proposed modified hierarchy shows improved classification results because  fewer examples are misclassified at higher levels, resulting in less error propagation. Comparison of our proposed approach with the competitive hierarchy modification approaches in the literature showed significant performance improvement supporting the hypothesis that our approach identifies the better set of inconsistent nodes. We also performed experiments to compare our approach with the flat approach with varying distribution of training examples per categories. Results demonstrated the usefulness of leveraging hierarchical information for classifying classes with fewer training examples. 

In future, we plan to study the effect of our modified hierarchy on various state-of-the-art HC approaches \cite{gopal2013recursive}.

\section*{Acknowledgment}
This work was supported by NSF grant \#202882 to HR.

\renewcommand{\bibfont}{\footnotesize}
\bibliographystyle{./IEEEtranBST/IEEEtran}
\bibliography{./IEEEtranBST/IEEEabrv,TopDown_INF_references}

\end{document}